\documentclass[letterpaper]{article} 
\usepackage{aaai25}  
\usepackage{times}  
\usepackage{helvet}  
\usepackage{courier}  
\usepackage[hyphens]{url}  
\usepackage{amsmath}
\usepackage{multirow}
\usepackage{amssymb}
\usepackage[capitalize,noabbrev]{cleveref}
\usepackage{graphicx} 
\usepackage{natbib}  
\usepackage{caption} 
\usepackage{algorithm}
\usepackage{algorithmic}

\usepackage{multirow}
\usepackage{amsmath}
\usepackage{amssymb}
\usepackage[capitalize,noabbrev]{cleveref}
\usepackage{subcaption}
\usepackage{makecell}

\usepackage{newfloat}
\usepackage{listings}
\DeclareCaptionStyle{ruled}{labelfont=normalfont,labelsep=colon,strut=off} 
\floatstyle{ruled}
\newfloat{listing}{tb}{lst}{}
\floatname{listing}{Listing}
\usepackage{newfloat}
\usepackage{listings}
\DeclareCaptionStyle{ruled}{labelfont=normalfont,labelsep=colon,strut=off} 
\floatstyle{ruled}
\newfloat{listing}{tb}{lst}{}
\floatname{listing}{Listing}

\title{Extract Free Dense Misalignment from CLIP}
\author{
    JeongYeon Nam\textsuperscript{\rm 1}\footnote{Corresponding author: jy.nam@navercorp.com},
    Jinbae Im\textsuperscript{\rm 1},
    Wonjae Kim\textsuperscript{\rm 2},
    Taeho Kil\textsuperscript{\rm 1}
}
\affiliations{
    \textsuperscript{\rm 1} NAVER Cloud AI 
    \textsuperscript{\rm 2} NAVER AI Lab
}

\usepackage{bibentry}

\begin{document}
\maketitle

\section{Abstract}
Recent vision-language foundation models still frequently produce outputs misaligned with their inputs, evidenced by object hallucination in captioning and prompt misalignment in the text-to-image generation model. Recent studies have explored methods for identifying misaligned elements, aiming not only to enhance interpretability but also to improve model performance. However, current approaches primarily rely on large foundation models in a zero-shot manner or fine-tuned models with human annotations, which limits scalability due to significant computational costs. This work proposes a novel approach, dubbed CLIP4DM, for detecting dense misalignments from pre-trained CLIP, specifically focusing on pinpointing misaligned words between image and text. We carefully revamp the gradient-based attribution computation method, enabling negative gradient of individual text tokens to indicate misalignment. We also propose F-CLIPScore, which aggregates misaligned attributions with a global alignment score. We evaluate our method on various dense misalignment detection benchmarks, covering various image and text domains and misalignment types. Our method demonstrates state-of-the-art performance among zero-shot models and competitive performance with fine-tuned models while maintaining superior efficiency. Our qualitative examples show that our method has a unique strength to detect entity-level objects, intangible objects, and attributes that can not be easily detected for existing works. We conduct ablation studies and analyses to highlight the strengths and limitations of our approach. Our code is publicly available at https://github.com/naver-ai/CLIP4DM.

\section{Introduction}

While recent advancements in generative models have garnered unprecedented progress, large-scale models still produce outputs misaligned with their inputs, exemplified by object hallucination~\cite{li2023evaluating,gunjal2024detecting} in image-to-text (captioning) models and misalignment with text description~\cite{rassin2022dalle,chefer2023attend} in text-to-image generation models. It is crucial to effectively detect these misalignments in order to develop a more reliable system.

\begin{figure}[t!]
     \centering
     \includegraphics[width=1.0\linewidth]{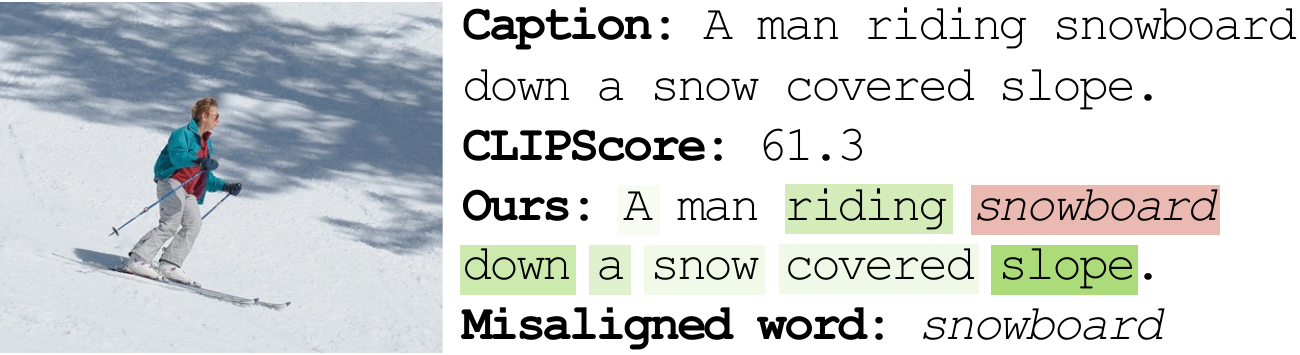}
     \caption{\textbf{Overview of our work.} CLIPScore indicates the alignment between the image and text in a single scalar score, limiting the interpretation of the score. Our approach extracts both positive and negative attributions to identify misaligned tokens between the image and text caption.}
     \label{fig:motivation}
     \vspace{-4mm}
\end{figure}

To measure the alignment between an image and text, the similarity score from CLIP \cite{radford2021learning,hessel2021clipscore} has become a de facto approach.
However, as shown in \cref{fig:motivation}, this simple score lacks the granularity needed to identify specific misaligned words, limiting interpretability~\cite{hu2023tifa,chodavidsonian}.
To address this limitation, recent studies \cite{petryk2024aloha,gordon2023mismatch} have focused on detecting misalignments at a dense level (e.g., word, phrase) and provide feedback to the models~\cite{yu2022scaling,yan2024vigor} for further enhancement. 
These approaches either employ a pipeline comprising multiple foundation models in zero-shot or fine-tuned configurations, leveraging costly human-annotated data.
While these methods show promising results, their computational expense limits their applicability in practical scenarios.

In this paper, we introduce a novel approach that leverages pre-trained CLIP for detecting dense misalignments efficiently. Specifically, our work aims at pinpointing words inconsistent with the image, offering richer explanations for text-image misalignments. 
While CLIP's final output is a single similarity score, we hypothesize that rich token-specific information is embedded within the model's intermediate representations, such as attention maps and gradients with respect to them.
We propose a new method, dubbed as CLIP4DM(\textbf{CLIP} for \textbf{d}ense \textbf{m}isalignment), which carefully modifies existing gradient-based attribution assignment techniques~\cite{selvaraju2017grad,chefer2021generic,chefer2021transformer}.
We compute attribution scores for each text token primarily based on relevance propagation methods, where our method is modified so that each relevance score can also be a negative attribution value. 
Then, we predict misaligned tokens by identifying text tokens with negative attribution lower than the threshold as shown in~\cref{fig:motivation}.
We also introduce F-CLIPScore, which combines the overall score with calculated attributions of misaligned tokens.

We thoroughly evaluate our method on diverse dense misalignment detection benchmarks (FOIL~\cite{shekhar2017foil}, nocaps-FOIL~\cite{petryk2024aloha}, HAT~\cite{petryk2024aloha}, SeeTRUE-Feedback~\cite{gordon2023mismatch}, and Rich-HF~\cite{liang2024rich}), encompassing various text, image, and misalignment types. The results consistently demonstrate that our method achieves state-of-the-art performance among zero-shot models and competitive performance with fine-tuned models.
Qualitative assessments reveal that our method robustly handles various misalignments, such as entity-level object class, intangible objects, and attributes. Moreover, our method demonstrates significantly higher efficiency compared to baselines, which utilize large foundation models, suggesting its potential for practical applications.

\section{Related Work}

\subsection{Dense Misalignment Detection}

There has been a growing emphasis on detecting dense misalignments between image and text, which focuses on identifying specific misaligned regions or tokens within the text.
This approach provides detailed feedback that improves the evaluation of image-text alignment. 
\citet{shekhar2017foil} introduce FOIL benchmark for detecting and correcting misaligned words, where replacing one noun of COCO caption with a semantically similar one.
ALOHa~\cite{petryk2024aloha} extends its coverage to various objects while leveraging multiple foundation models.
ALOHa makes a candidate object pool with an extracted noun phrase from reference captions and the results of object detectors~\cite{carion2020end}, then perform bipartite matching based on scores derived from a language semantic similarity model~\cite{reimers2019sentence}.
SeeTRUE-Feedback~\cite{gordon2023mismatch} utilizes LLMs and a visual grounding model to create a dataset of textual and visual misalignment descriptions, which are then used to train a vision-language model for automatic feedback generation.
Rich-HF~\cite{liang2024rich} focuses on misalignments in a text-to-image generation model while collecting human annotations on misaligned keywords and implausible image regions and trains a multimodal language model to show the dense image-text alignment automatically.

Beyond simply detecting dense misalignments, there have been studies leveraging dense misalignment labels to enhance model performance or reduce object hallucinations, particularly in the context of reinforcement learning-based approaches. 
As the length of sequences generated by LLMs increases, the problem of hallucination becomes pronounced, making dense feedback that reduces ambiguity inherent in single scalar reward more critical.
\citet{yu2024rlhf} and \citet{xiao2024detecting} tackle object hallucination in large vision-language models by incorporating dense-level (e.g., sub-sentence, sentence) human feedbacks.
ViGoR~\cite{yan2024vigor} additionally employs a pipeline combining named entity recognition models with open vocab object detector~\cite{liu2023grounding} to detect hallucinations automatically. However, its scope is limited to object hallucinations, and human annotations are still needed to detect comprehensive misalignments.

In summary, the increasing emphasis on dense misalignment detection underscores its crucial role in developing more interpretable and reliable vision-language models. 
While current work demonstrates promising results in providing dense misalignment detection, they predominantly rely on costly human annotations or incorporation of foundation models, resulting in substantial cost overhead.
In this work, we propose a cost-efficient dense misalignment detection method, leveraging the pre-trained CLIP in a zero-shot manner.
The result demonstrates its efficiency and competitive performance over other cost-expensive zero-shot baselines.

\subsection{Explainable AI Methods}
Understanding the decision-making process of complex machine learning models is crucial for building trust and ensuring reliable performance.
Explainable AI (XAI) methodologies address this need by providing insights into how models arrive at their predictions.
XAI methods can be broadly categorized into two groups: input manipulation methods and mechanistic approaches.
Input manipulation methods, such as SHAP~\cite{lundberg2017unified}, occlusion analysis~\cite{zeiler2014visualizing}, and LIME~\cite{ribeiro2016should}, perturb or mask input features to observe their impact on model output.
While intuitive, these methods are often computationally expensive, especially for large models and datasets.

Mechanistic approaches, on the other hand, delve into the internal workings of the model to directly analyze feature contributions.
Grad-CAM~\cite{selvaraju2017grad} uses class-specific gradients to highlight relevant input regions but can produce coarse visualizations.
LRP~\cite{bach2015pixel}, grounded in the Deep Taylor Decomposition framework~\cite{montavon2017explaining}, propagates relevance scores backward through the network layers, ensuring conservation of relevance.
LRP has been successfully applied to various tasks, including image classification~\cite{bach2015pixel}, NLP~\cite{arras2017explaining}, and vision-and-language tasks~\cite{chefer2021transformer}, showcasing its versatility and effectiveness.

The widespread adoption of Transformer networks~\cite{vaswani2017attention} in NLP and vision-and-language tasks brought new challenges for XAI.
Rollout~\cite{abnar2020quantifying} and Attention Flow~\cite{abnar2020quantifying} attempt to address complexities arising from self-attention, but limitations persist.
\citet{chefer2021transformer} adapted LRP for single-modality Transformers, later extending it to multi-modal settings using a combination of attention scores and gradients for head averaging~\cite{chefer2021generic}.
Unlike these approaches, which rely on positive-only relevance propagation, our work introduces the interpretation of negative attributions as indicators of misalignment in CLIP.
Recent studies~\cite{zhou2022extract,wang2023visual,zhao2024gradient} apply XAI techniques to CLIP; however, they also focus on identifying only relevant image regions corresponding to the text.

\section{Method}

\subsection{Preliminary: CLIP}
We provide a brief overview of the key elements of the CLIP architecture.
We also define the relevant terminology to consistently notate our method.

CLIP employs a dual-encoder structure, processing image and text modalities through separate encoders. 
The text encoder takes a sequence of tokens padded or truncated to a fixed length $n$,
\begin{equation}
t = [t_0, t_1, ..., t_{z}, ..., t_{n-1}],
\end{equation}
where $z$ is the index of the [EOS] token in the sequence.
The image encoder processes the input image as a sequence of patches, including a special [CLS] token.
\begin{equation}
v = [v_0, v_1, ..., v_m],
\end{equation}
where $v_0$ is the [CLS] token and $v_1, ..., v_m$ are image patches.
The input image patches $v$ and text tokens $t$ are first forwarded through the image encoder ($V$) and text encoder ($T$), respectively, after which the representations are pooled and projected from the [CLS] and [EOS] tokens:
\begin{equation}
e_v = W_v(V(v)[0, :]), \quad e_t = W_t(T(t)[z, :]),
\end{equation}
where $W_v$ and $W_t$ are projection matrices.
The final score is computed by the cosine similarity (dot product with L2 normalization):
\begin{equation}
\text{score}_{v, t} = \frac{e_v}{||e_v||_2} \cdot \frac{e_t}{||e_t||_2}.
\end{equation}
This score indicates the degree of semantic alignment between the image and text inputs.
\begin{table*}[t]
\small
\centering
\setlength{\tabcolsep}{0.4\tabcolsep}
\begin{tabular}{cccccccc}
\hline
Benchmark & Source & Text / Image & Misalign & Num of & Annotation & Dense & Global \\
& & domain & Type & Misaligns & Type& Misalign & Misalign \\
\hline
FOIL & COCO caption & natural / natural & object & single & rule-based & Accuracy & Average Precision \\
\hline
nocaps-FOIL & nocaps & natural / natural & object & single & rule-based &  Accuracy & Average Precision \\
\hline
HAT & COCO caption & generated / natural & various & multiple & human &  Accuracy & Average Precision \\
\hline
\multirow{4}{*}{SeeTRUE-Feedback} &  COCO-con & natural / natural & various & multiple & human& NLI Score & - \\
 & COCO-T2I & natural / generated & various & multiple & human  & NLI Score & - \\
& Drawbench & natural / generated & various & multiple & human  & NLI Score & - \\
& Pick-a-pic-con & generated / generated & various & multiple & human & NLI Score & - \\
\hline
Rich-HF & Pick-a-pic & natural / generated & various & multiple & human & precision, recall, F1 & corr. coeff. \\
\hline
\end{tabular}
\caption{\textbf{Comprehensive overview of benchmarks for dense misalignment detection.} ``Generated'' in the Text / Image domain column indicates that the text or image was created by a captioning model or a text-to-image generation model, respectively. In contrast, ``natural'' signifies that the text or image originates from a human source.
}
\label{tab:benchmarks-revised}
\end{table*}

\subsection{Our Method}
In this section, we introduce our attribution calculation method, which is inspired by Generic Attention-model Explainability (GAE)~\cite{chefer2021generic}.
We first introduce GAE briefly and how our method is different from GAE.
We then introduce fine-grained CLIPScore (F-CLIPScore), a drop-in replacement of CLIPScore by aggregating word attributions.
\subsubsection{Generic Attention-model Explainability}
To determine the direction and magnitude of each token's attribution to the final output, GAE computes the gradients of the final score with respect to the attention map:
\begin{equation}
\nabla A_l^h = \frac{\partial \text{score}_{v, t}}{\partial A_l^h},
\end{equation}
where $A_l^h \in \mathbb{R}^{n \times n}$ denotes the attention map at $l$-th layer and $h$-th head.
To aggregate its gradient, GAE calculates the element-wise product of this gradient with the corresponding attention map:
\begin{equation}
R_l^h = \text{ReLU}(\nabla A_l^h \odot A_l^h).
\label{eq:relu}
\end{equation}
Note that relevance propagation methods~\cite{selvaraju2017grad,chefer2021generic,chefer2021transformer,montavon2017explaining} typically employ ReLU operation in $\nabla A_l^h$ to remove negative attribution. 

\begin{table*}[t!]
\small
\centering
    \setlength{\tabcolsep}{0.5\tabcolsep}
    \begin{center}
        \begin{tabular*}{\linewidth}{@{\extracolsep{\fill}} lcccccccccccc}
        \hline
            \multirow{3}{*}{Method} & \multirow{3}{*}{FPS} & \multirow{3}{*}{\shortstack{needs\\annotations}} & \multicolumn{2}{c}{\multirow{2}{*}{FOIL}} & \multicolumn{8}{c}{nocaps-FOIL} \\
            \cline{6-13}
            & & & & & \multicolumn{2}{c}{Overall} & \multicolumn{2}{c}{In-Domain} & \multicolumn{2}{c}{Near-Domain} & \multicolumn{2}{c}{Out-of-Domain} \\
            \cline{4-5} \cline{6-13}
            & & & LA & AP & LA & AP & LA & AP & LA & AP & LA & AP \\
            \hline
            CHAIR & - & \checkmark & \underline{0.790} & \textbf{0.925} & 0.144 & 0.583 & 0.135 & 0.578 & 0.176 & 0.591 & 0.122 & 0.581 \\
            CLIPScore (ViT-B/32) & 13.4 &  & - & \textit{0.707} & - & \textit{0.692} & - & \textit{0.651} & - & \textit{0.675} & - & \textit{0.743} \\
            CLIPScore (ViT-H/14) & 8.72  & & - & 0.763 & - & 0.722 & - & 0.690 & - & 0.707 & - & 0.764 \\
            RefCLIPScore (ViT-B/32) & 8.75 & \checkmark & - & \textit{0.748} & - & \underline{\textit{0.736}} & - & \textit{0.683} & - & \underline{\textit{0.718}} & - & \underline{\textit{0.791}} \\
            ALOHa & 0.16 & \checkmark & 0.400 & 0.614 & 0.452 & 0.695 & 0.474 & \underline{0.718} & 0.473 & 0.667 & 0.488 & 0.709 \\
            \hline
            Ours (ViT-B/32) & 12.0 & & 0.732 & 0.714 & \underline{0.603} & 0.690 & \underline{0.547} & 0.673 & \underline{0.597} & 0.684 & \underline{0.632} & 0.713 \\
            Ours (ViT-H/14) & 7.06 & & \textbf{0.836} & \underline{0.806} & \textbf{0.716} & \textbf{0.794} & \textbf{0.661} & \textbf{0.789} & \textbf{0.708} & \textbf{0.793} & \textbf{0.748} & \textbf{0.802} \\
            \hline
        \end{tabular*}
    \end{center}
    \vspace{-1.0em}
    \caption{\textbf{Experiment results on FOIL and nocaps-FOIL}. LA: Localization Accuracy. AP: Average Precision. FPS is measured on the nocaps-FOIL dataset. \textit{Italic} denotes that we remeasured the result with ViT-B/32. }
    \vspace{-2.mm}
    \label{table:foil}
\end{table*}

The relevancy for layer $l$ is obtained by averaging across attention heads:
\begin{equation}
R_l = \frac{1}{H} \sum_{h=1}^{H} R_l^h.
\label{eq:headmean}
\end{equation}
The relevancy in the final layer is initialized as an identity matrix and updated layer by layer. At each layer $l$, $R$ is updated by adding the product of the current layer's relevancy $R_l$ and the carried $R$ as in the relevance propagation methods.
This process propagates the attribution information through the network, accumulating each layer's attribution.
Finally, the relevancy is aggregated along the [EOS] token row, $R[z,:]$.

\subsubsection{Allowing Negative Gradient Flow.} 
Unlike GAE~\cite{chefer2021generic}, which focuses on only the positive value of gradient, our work aims to identify misaligned words by incorporating negative gradients. We simply remove the ReLU operation on \cref{eq:relu}, allowing negative gradients to explain the model's behavior.
\begin{equation}
R_l^h = \nabla A_l^h \odot A_l^h.
\end{equation}
By adopting this formulation, our approach leverages both positive and negative gradients to capture a comprehensive spectrum of attributions.

\subsubsection{Layer Aggregation.}
Since our method incorporates gradients of both signs, matrix multiplication could lead to ambiguous interpretations.
To address this, we average the attribution map $R_l$ across layers, preserving the interpretability of both positive and negative attributions.
\begin{equation}
R = \frac{1}{(L - \tilde{l} + 1)} \sum_{l=\tilde{l}}^{L} R_l,
\end{equation}
where $L$ is the total number of layers in the transformer model, $\tilde{l}$ is the index of the starting layer for accumulation, which is a hyperparameter. 
\subsubsection{Token Aggregation and F-CLIPScore.} 
To identify misaligned words, we calculate the word-level attribution $w_j$ by averaging the attribution of its constituent tokens. We then predict a word as misaligned if its attribution falls below a threshold $\epsilon$. 
\begin{equation}
\text{mis}(w_j) = \begin{cases}
1, & \text{if } w_j < \epsilon \\
0, & \text{otherwise.}
\end{cases}
\end{equation}
To get a global fine-grained misalignment score between images and text, similar to CLIPScore~\cite{hessel2021clipscore}, we devise a simple aggregation method to derive a single score, which is dubbed as F-CLIPScore, as follows:
\begin{equation}
\text{F-CLIPScore}(v, t) = (1-\text{score}_{v,t})\cdot\sum_j\text{mis}(w_j) \cdot w_j .
\label{eq:score}
\end{equation}

This aggregation integrates both overall semantic alignment and fine-grained misalignments for each token.

\section{Experiments}
As summarized in \cref{tab:benchmarks-revised}, we comprehensively evaluate our method across a diverse range of dense misalignment detection benchmarks. Our evaluation spans text domains (natural and generated), image domains (natural and generated), misalignment types (object, attribute, relation, and action), and the number of misaligned words (single or multiple). This extensive testing demonstrates the robustness and versatility of our approach. For detailed information about the datasets and experiments on additional benchmarks, please refer to the supplementary materials.

We report two variants of CLIP: OpenAI CLIP ViT-B/32~\cite{radford2021learning}, following 
~\citet{hessel2021clipscore}, and ViT-H/14 trained on LAION-2B~\cite{schuhmann2022laion} from OpenClip~\cite{cherti2023reproducible}, which yields our best score. Further analysis of other backbones is provided in the supplementary material. We use a template ``A photo depicts '' following ~\citet{hessel2021clipscore}.
We set our hyperparameters by searching the development set of Rich-HF and a subset of the training set from the FOIL dataset. We use $\tilde{l}$ to 10 and 22 for ViT-B/32 and ViT-H/14, respectively, utilizing the final three layers in both cases. Unless otherwise specified, $\epsilon$ is set to -0.00005. Frames-Per-Second (FPS) is measured with a single V100. Finally, we use F-CLIPScore for the global misalignment classification task. 

\subsection{Quantitative Results}

\subsubsection{FOIL and nocaps-FOIL.}
FOIL~\cite{shekhar2017foil} and nocaps-FOIL~\cite{petryk2024aloha} are benchmarks for detecting misaligned captions where one object is replaced by a conceptually similar word (e.g., car, bicycle).
We assess performance on two protocols: (1) localization accuracy (LA) for dense misalignment detection and (2) average precision (AP) for global misalignment classification. Following existing works, our approach predicts a single word with the lowest attribution $w_j$. 
In nocaps-FOIL, we report results as in-domain, near-domain, or out-of-domain based on how similar the altered objects are to COCO object classes.

In~\cref{table:foil}, our ViT-B/32 variant demonstrates state-of-the-art performance on most dense misalignment detection (LA). It is worth noting that baselines such as CHAIR~\cite{rohrbach2018object} or ALOHa~\cite{petryk2024aloha} make use of ground truth segmentation labels or reference captions. 
The ViT-H/14 variant demonstrates significantly enhanced performance, showing improved results consistently across all domains. It shows the robustness of our approach, which utilizes CLIP model pre-trained on various alt-text. 
Furthermore, our F-CLIPScore boosts up global misalignment classification (AP) by a significant margin, even surpassing reference-based methods. 
Lastly, our proposed approach demonstrates significantly superior computational efficiency compared to ALOHa, achieving a 44-fold reduction in inference time.

\subsubsection{HAT.}
The HAT dataset~\cite{petryk2024aloha} comprises 400 human-annotated samples featuring captions generated by VLM models~\cite{li2022blip,wang2022ofa,chan2023ic3,zhu2023chatgpt}. 
For evaluation, we measured with the same metric as FOILs: LA and AP. For LA, correctly identifying any hallucinated object in a sentence is considered accurate. To compare with ALOHa, which extracts a noun phrase, we concatenate neighboring misaligned words and average scores within a phrase. The phrase with the lowest aggregate score was predicted as the erroneous segment.

In~\cref{table:hat}, our ViT-H/14 variant demonstrates superior performance in LA with significantly improved FPS. 
In terms of AP, our method shows a performance gap compared to models that utilize reference captions. Further analysis for AP is presented in the supplementary.
\begin{table}[ht!]
\small
\centering
\setlength{\tabcolsep}{0.5\tabcolsep}
\begin{center}
\begin{tabular*}{\linewidth}{@{\extracolsep{\fill}} lcccc}
\hline
method & \shortstack{ref. captions} & FPS & LA & AP \\
\hline
CHAIR &  \checkmark& - & 0.067 & 0.369 \\
CLIPScore (ViT-B/32)  & & 18.8 & - & \textit{0.385} \\
RefCLIPScore (ViT-B/32)   & \checkmark & 9.03 & - & \underline{\textit{0.429}} \\
ALOHa &   \checkmark & 0.24 & \underline{0.203} & \textbf{0.486} \\
\hline
Ours (ViT-B/32) & &9.64 & 0.193 & 0.355 \\
Ours (ViT-H/14)  & & 6.56 & \textbf{0.348} & 0.360 \\
\hline
\end{tabular*}
\end{center}
\vspace{-1.0em}
\caption{\textbf{Experiment results on HAT test set.} ref. captions denotes that the method utilizes reference captions. \textit{Italic} denotes that we remeasured the result with ViT-B/32.}
\vspace{-2.mm}
\label{table:hat}
\end{table}
\begin{table}[t!]
\centering
\setlength{\tabcolsep}{0.5\tabcolsep}
\begin{center}
\begin{tabular*}{\linewidth}{@{\extracolsep{\fill}} lccc}
\hline
Model & ft. & FPS & NLI score \\
\hline
LLaVa-1.5 (Vicuna-7B)&  & 0.24 &0.173 \\
PaLI 5B & & - & 0.226 \\
mPLUG-Owl (LLaMa-7B) & & 0.24 & 0.297 \\         
InstructBLIP (FlanT5$_{\text{XL}}$) &  & 0.51 & 0.555 \\
MiniGPT-v2 (LLaMa2-7B) &  & 0.28 &0.560 \\
\hline
Ours (ViT-B/32) & & 7.90 & 0.605 \\
Ours (ViT-H/14) & & 5.81 &0.660 \\
\hline
PaLI 5B  & \checkmark & - & \underline{0.765} \\
PaLI 17B  & \checkmark & - & \textbf{0.785} \\
\hline
\end{tabular*}
\end{center}
\vspace{-1.0em}
\caption{\textbf{Textual misalignment performance on SeeTRUE-Feedback test set.} ft. denotes that the model is fine-tuned.}
\vspace{-2.mm}
\label{table:seetrue}
\end{table}

\subsubsection{SeeTRUE-Feedback.}
SeeTRUE-Feedback~\cite{gordon2023mismatch} comprises a test set of 2K samples covering various images, text domains, and misalignment types.
We specifically focus on textual misalignment detection, which aims to extract mismatched spans from caption.
Following established protocols, we report the natural language inference (NLI)~\cite{bowman2015large} score obtained from a BART-NLI model~\cite{lewis2020bart}. We calculate the entailment score where the premise is the ground truth label, and the hypothesis is the predicted word span. To form a single sequence, we take the same strategy as the one used in evaluating the HAT dataset.

As shown in~\cref{table:seetrue}, our method surpasses the zero-shot models, showing its efficiency and robustness in various domains. 
It is also worth noting that, as a non-generative model, ours offers faster inference times compared to larger vision language models.

\subsubsection{Rich-HF.}
Rich-HF~\cite{liang2024rich} comprises 955 prompt and image pairs with word-level misalignment annotations and overall alignment score. Since prompts are collected by real users~\cite{kirstain2023pick}, its captions cover various lengths, styles, and contents. We evaluate the performance of misalignment labels using precision, recall, and F1 scores at the word level. We also measure Pearson and Spearman's correlation between our aggregated score and the Likert score for alignment. We further report the performance of ALOHa for comparison.

In~\cref{table:richt2}, our method demonstrates promising performance as a zero-shot method. While precision is limited, it shows higher recall, resulting in a substantial F1 score.
In~\cref{table:richt1}, our aggregated score shows a significantly enhanced score in two correlation coefficients.
Our ViT-H/14 variant even shows comparative performance with PickScore~\cite{kirstain2023pick}, which is finetuned with 583K human preference scores. These results suggest that the selected negative attributions effectively capture misalignment, leading to superior performance in measuring text-image discrepancies.

\begin{table}[t!]
\centering
    \setlength{\tabcolsep}{0.5\tabcolsep}
    \begin{center}
        \begin{tabular*}{\linewidth}{@{\extracolsep{\fill}} lcccc}
        \hline
            Model & ft. &  F1 & precision & recall  \\
            \hline
            ALOHa & & 0.344 & 0.311 & 0.385\\
            \hline
            Ours (ViT-B/32)$_{\epsilon=-0.00001}$ & & 0.398 & 0.328 & \underline{0.504}  \\
            Ours (ViT-H/14)$_{\epsilon=-0.00001}$ & & 0.427 & 0.365 & \textbf{0.516} \\
            Ours (ViT-H/14)$_{\epsilon=-0.00005}$ & & 0.314 & 0.487 & 0.231   \\
            \hline
            Rich-HF (multi-head) & \checkmark &  \underline{0.433} & \textbf{0.629} & 0.330  \\
            Rich-HF (augmented prompt) & \checkmark & \textbf{0.439} & \underline{0.613} & 0.341  \\
            \hline
        \end{tabular*}
    \end{center}
    \vspace{-1.0em}
    \caption{\textbf{Experiment results on Rich-HF test set}. ft. denotes that the model is fine-tuned with the Rich-HF training set.}
    \vspace{-2.mm}
    \label{table:richt2}
\end{table}

\begin{table}[t!]
\centering
    \setlength{\tabcolsep}{0.5\tabcolsep}
    \begin{center}
        \begin{tabular*}{\linewidth}{@{\extracolsep{\fill}} lcccc}
        \hline
            Model & ft. & pearson & spearman \\
            \hline
            CLIPScore (ViT-B/32) & & 0.185 & 0.130 \\
            PickScore (ViT-H/14) &  & 0.346 & 0.340 \\
            \hline
            Ours (ViT-B/32)$_{\epsilon=-0.00001}$ & & 0.279 & 0.332\\
            Ours (ViT-H/14)$_{\epsilon=-0.00001}$ & & 0.368& 0.433 \\
            \hline
            CLIPScore (ViT-B/32) & \checkmark & 0.398 & 0.390 \\
            Rich-HF (multi-head) & \checkmark & \textbf{0.487} & \textbf{0.500} \\
            Rich-HF (augmented prompt) & \checkmark & \underline{0.474} & \underline{0.496} \\
            \hline
        \end{tabular*}
    \end{center}
    \vspace{-1.0em}
    \caption{\textbf{Experiment results on Rich-HF misalignment score correlation.}}
    \vspace{-2.mm}
    \label{table:richt1}
\end{table}

\subsection{Qualitative Results}
We present qualitative examples on three representative datasets in \cref{fig:oursqual}. Further examples of all datasets are shown in the supplementary materials.
For the FOIL and nocaps-FOIL datasets, models need to predict a single word regardless of the presence or absence of misaligned words. 
When misaligned words exist, our model detects them well for images from various domains. 
In cases where misaligned words do not exist, our model predicts unimportant word `.' and `medium', a word that is difficult for the model to distinguish, as misaligned words as shown in the second and fourth images for the FOIL dataset.
For the Rich-HF dataset, our model demonstrates decent misaligned word detection performance for generated images. In addition, ours shows the ability to detect multiple misaligned words or not detect misaligned words when misaligned words do not exist.

\begin{figure*}[t]
     \centering
     \includegraphics[width=1.0\linewidth]{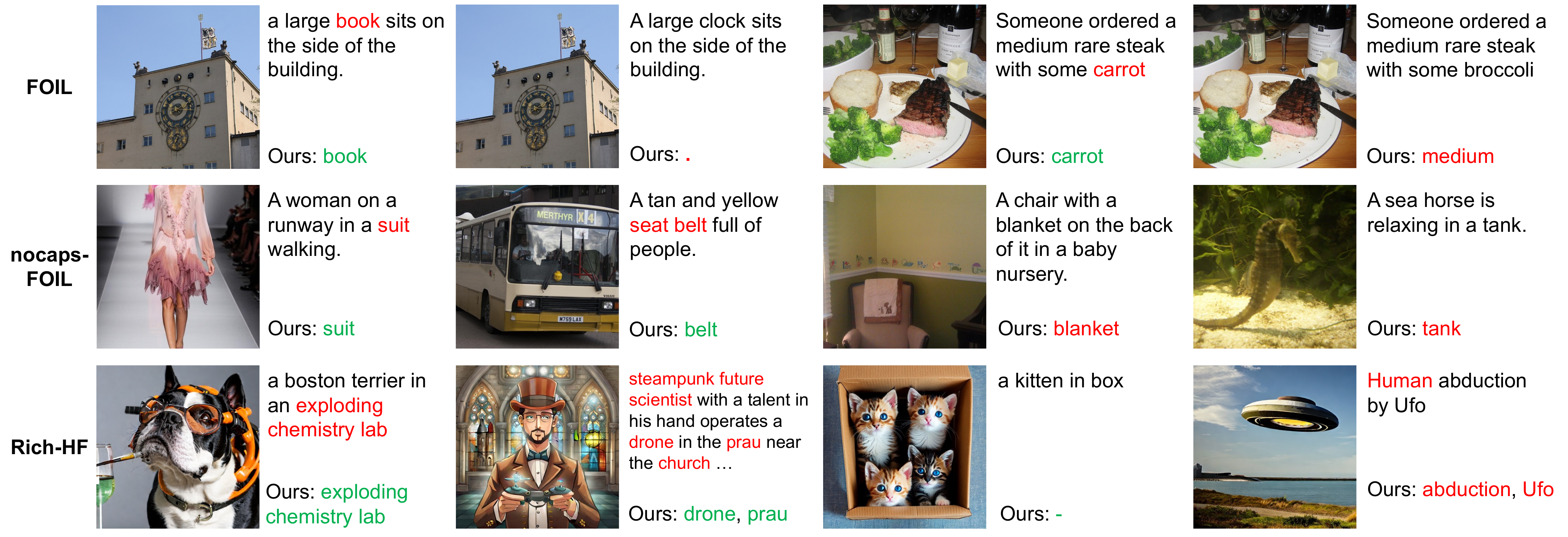}
     \caption{\textbf{Qualitative examples on FOIL, nocaps-FOIL, and Rich-HF datasets.} Misaligned words are highlighted in red in captions paired with images. Note that misaligned words may not exist. For predicted misaligned words, correct words are shown in green and incorrect words in red. If our model predicts that there are no misaligned words, it is indicated as `-'. }
     \label{fig:oursqual}
     \vspace{-4mm}
\end{figure*}

\subsection{Ablation Studies}
To demonstrate the efficiency of the proposed method, we conduct ablation studies using the ViT-H/14 variant.

\begin{table}[t!]
\centering
    \setlength{\tabcolsep}{0.5\tabcolsep}
    \begin{center}
        \begin{tabular*}{\linewidth}{@{\extracolsep{\fill}} lccc}
        \hline \\
        Method & FPS & LA&  AP  \\
        \hline
        occlusion-based & 0.6& 0.566 & 0.748  \\
        \hline
        gradient-based \\
        $\nabla A_l^h$ & 5.8 & 0.423& 0.741  \\
        $A_l^h \odot \nabla  A_l^h$ & 5.8 & \textbf{0.716}&  \textbf{0.794}  \\
        \hline
        \end{tabular*}
    \end{center}
    \vspace{-1.0em}
    \caption{\textbf{Ablation study of attribution calculation methods on the nocaps-FOIL test set.} }
    \vspace{-2.mm}
    \label{table:ablation-method}
\end{table}

\begin{table}[t!]
\centering
    \setlength{\tabcolsep}{0.5\tabcolsep}
    \begin{center}
        \begin{tabular*}{\linewidth}{@{\extracolsep{\fill}} cccc}
        \hline
            $ReLU(-\nabla A_l^h)$ & $ReLU(-\nabla A_l)$& LA  &  AP \\
            \hline
            \checkmark & & 0.698& 0.779  \\
            & \checkmark & 0.700& 0.776  \\
            && \textbf{0.716}& \textbf{0.794}  \\
            \hline
        \end{tabular*}
    \end{center}
    \vspace{-1.0em}
    \caption{\textbf{Ablation study of the disabling positive gradients on nocaps-FOIL test set.} ReLU($-\nabla A^h_l$) and ReLU($-\nabla A_l$) indicate retaining only negative gradients before averaging across heads and layers, respectively. 
    }
    \vspace{-2.mm}
    \label{table:ablation-relu}
\end{table}

\begin{figure}[t]
     \centering
     \includegraphics[width=1.0\linewidth]{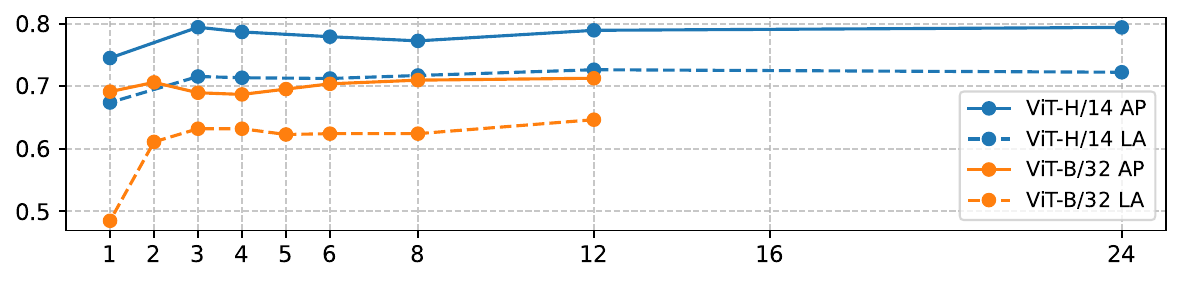}
     \caption{\textbf{Ablation on the number of text encoder layers used for attribution calculation on nocaps-FOIL dataset.}}
     \label{fig:layer}
     \vspace{-4mm}
\end{figure}

\subsubsection{Attribution Calculation Method.}

We conduct an ablation study on the attribution calculation method. The occlusion~\cite{goyal2016towards} method iteratively omits individual words from the input text and identifies the word whose removal leads to the highest increase in the score as the most likely erroneous element. Among gradient-based methods, we ablate components used in extracting attribution maps. \cref{table:ablation-method} shows that the occlusion-based method demonstrated superior performance, but its efficiency was limited due to the requirement of multiple forward passes. In contrast, gradient-based methods, particularly when combined with attention maps, achieved a balance of high efficiency and performance. 

\subsubsection{Disabling Positive Gradient.}
We examine the effectiveness of using both positive and negative gradients in attribution calculation. We compare removing positive gradients before averaging across heads or layers, similar to conventional relevance map approaches~\cite{selvaraju2017grad,chefer2021generic}. \cref{table:ablation-relu} demonstrates that utilizing full gradients yields the best performance, outperforming methods that isolate negative gradients. 
This finding underscores the importance of considering both positive and negative contributions in gradient-based attribution calculation.

\subsubsection{Number of Layers.}
We perform an ablation study on $\tilde{l}$, the number of text encoder layers used for attribution map calculation. \cref{fig:layer} demonstrates that utilizing multiple layers, rather than solely the final layer, significantly enhances performance across both metrics. 
It demonstrates that utilizing intermediate features across layers enhances the detection of misalignments.

\subsubsection{Components of F-CLIPScore.}
\begin{table}[t!]
\centering
\footnotesize
\setlength{\tabcolsep}{0.5\tabcolsep}
\begin{tabular}{llccc}
\hline
Dataset & Method & AP & Pearson & Spearman  \\
\hline
\multirow{3}{*}{nocaps-FOIL} 
&$\text{score}_{v,t}$ & 0.722 &- & -\\
& $\sum_j\text{mis}(w_j) \cdot w_j$ & 0.776 & - & -\\
& F-CLIPScore  & \textbf{0.794} & - & -\\
\hline
\multirow{3}{*}{Rich-HF} 
& $\text{score}_{v,t}$ & - & 0.171 & 0.085  \\
& $\sum_j\text{mis}(w_j) \cdot w_j$  & - &0.352 & 0.419 \\
& F-CLIPScore  & - & \textbf{0.368} & \textbf{0.433} \\
\hline
\end{tabular}
\caption{\textbf{Ablation on components of  F-CLIPScore.} 
Result shows that simple aggregating negative attributions can enhance the capture of alignments, which is further improved with a combination of the overall score. }
\label{table:ablation-components}
\end{table}
We conduct an ablation study on the components of F-CLIPScore using the nocaps-FOIL and Rich-HF datasets. As shown in \cref{table:ablation-components}, the mere summation of negative attribution $w_j$ yields significantly improved AP and correlation coefficients. Moreover, integrating this with the overall similarity score further enhances performance, demonstrating our method's efficacy in capturing fine-grained misalignments. Additional analyses are presented in the supplementary material.

\subsection{Analysis}
\subsubsection{Comparsion with Baselines.}

\begin{figure}[t]
     \centering
     \includegraphics[width=1.0\linewidth]{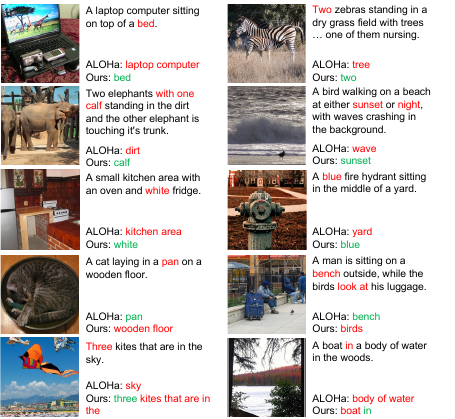}
     \caption{\textbf{Qualitative examples compared to ALOHa on HAT dataset.} Our method demonstrates improved robustness in various misalignment types.}
     \label{fig:qualaloha}
     \vspace{-4mm}
\end{figure}

In~\cref{fig:qualaloha}, we present qualitative examples comparing our method to the baseline ALOHa~\cite{petryk2024aloha} on the HAT dataset. Our approach demonstrates robust and diverse detection capabilities, such as colors (e.g., white), numbers (e.g., two), entity-level objects (e.g., calf), and intangible objects (e.g., sunset), which can not be easily captured with combinations of foundation models. 
Different from ALOHa, which uses a language similarity module, CLIP, which is trained on diverse alt-text data, is sensitive to conceptually similar but visually distinct words (e.g., ``blue'' and ``grey''). This underscores the effectiveness of our CLIP-based approach, which operates independently of additional foundation models. While showing promising results, our method also reveals some inherent limitations of CLIP, particularly in identifying discrepancies related to backgrounds (e.g., ``wooden floor'') or small objects (e.g., ``birds''). Further examples and analyses are presented in the supplementary materials.

\begin{figure}[t]
     \centering
     \includegraphics[width=1.0\linewidth]{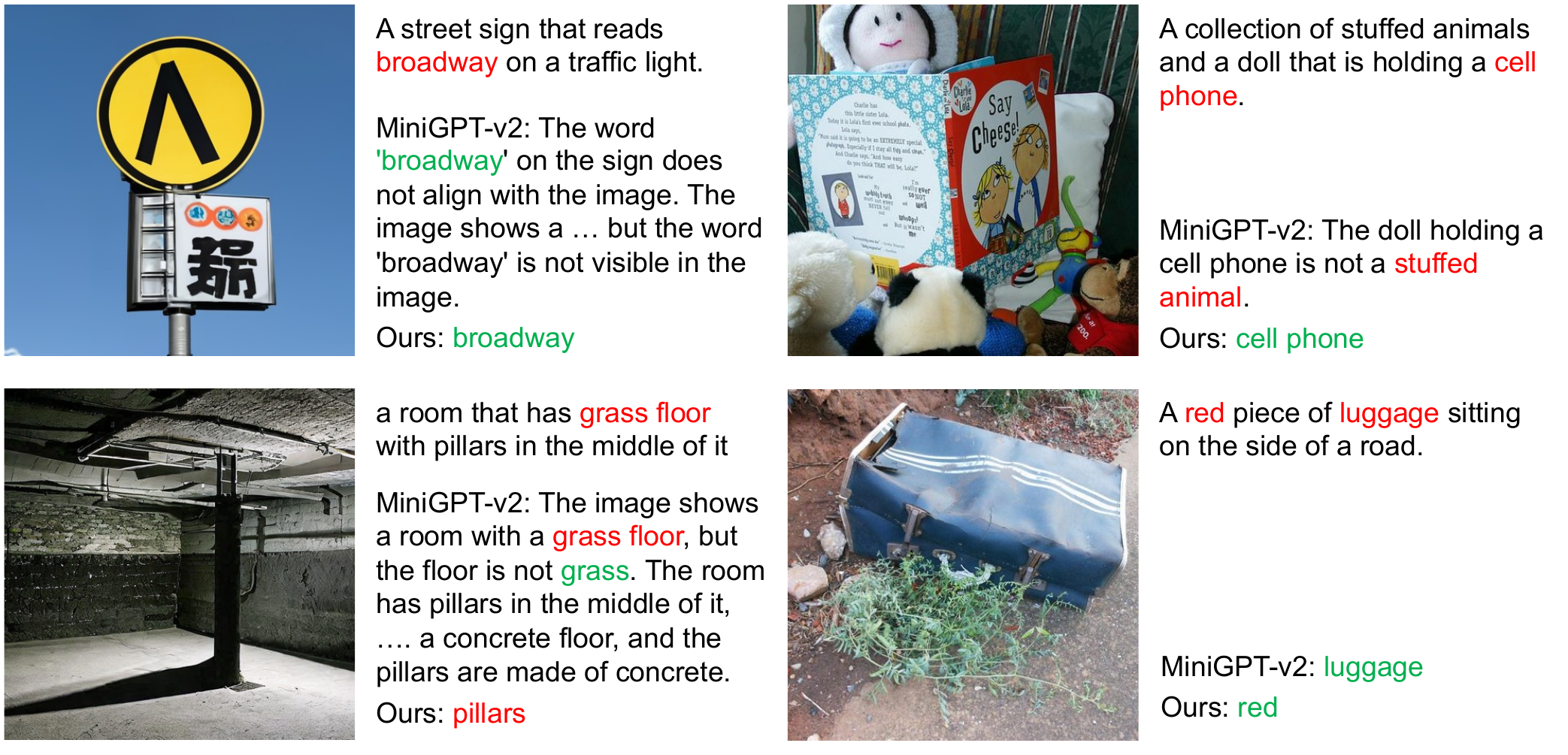}
     \caption{\textbf{Qualitative examples compared to MiniGPT-v2 on SeeTRUE-Feedback dataset.} MiniGPT-v2 generates lengthy and unformalized responses that are hard to parse into misaligned words for most examples. 
     }
     \label{fig:qualminigpt}
     \vspace{-4mm}
\end{figure}

We provide qualitative examples comparing our method to the baseline MiniGPT-v2~\cite{chen2023minigpt} on SeeTRUE-Feedback dataset, as shown in~\cref{fig:qualminigpt}.
As a large vision-language model, MiniGPT-v2 has the advantage of providing natural and rich responses. However, despite a prompt that requests the model to answer in short words, it provides lengthy and unformatted responses to almost all examples. 
Since it is quite difficult to accurately extract misaligned words from unstructured responses, its usability as a dense misalignment detector is low.
Furthermore, MiniGPT-v2 sometimes generates inconsistent responses.
As shown in the third example in~\cref{fig:qualminigpt}, it shows a contradictory response, saying that the floor is grass while also saying that it is not grass.
On the contrary, our method can detect misaligned words efficiently.
It is also noteworthy that the FPS of ours is far higher than that of MiniGPT-v2.

\subsubsection{Part-of-Speech.}

We report Rich-HF word level metrics per part-of-speech (POS) for further analysis. 
In~\cref{table:analysis-pos}, our method predicts all overall POS, which shows that our method has the capability to predict misaligned words of various types, not limited to nouns. Still, we observe a trend that shows decent performance with nouns but limited performance with adverbs, adjectives, numbers, and adpositions. The result shows that our result corresponds with studies that reveal CLIP's weaknesses~\cite{paiss2023teaching,nikolaus2022vision,yuksekgonul2023and}. We leave it as future work to test with CLIP variants, which are further fine-tuned to tackle such shortcomings.

\begin{table}[t!]
\small
\centering
\setlength{\tabcolsep}{0.5\tabcolsep}
\begin{center}
\begin{tabular*}{\linewidth}{@{\extracolsep{\fill}} lccccccc}
\hline
Metric & NOUN & PROPN & VERB & ADV & ADJ & NUM & ADP \\
\hline
F1 & 0.393 & 0.312 & 0.301 & 0.258 & 0.258 & 0.132 & 0.177 \\
Precision & 0.470 & 0.602 & 0.567 & 0.444 & 0.417 & 0.500 & 0.278 \\
Recall & 0.337 & 0.211 & 0.205 & 0.182 & 0.187 & 0.076 & 0.130 \\
\hline
\end{tabular*}
\end{center}
\vspace{-1.0em}
\caption{\textbf{Comparison of metrics per part-of-speech on Rich-HF test set.}}
\vspace{-2.mm}
\label{table:analysis-pos}
\end{table}
\section{Conclusion and Future Work}
In this paper, we present a novel approach for detecting dense misalignments between text and image using a pre-trained CLIP model. By extracting attributions from CLIP’s intermediate gradients, our method provides a scalable and efficient solution that achieves state-of-the-art performance in zero-shot settings and competitive results with fine-tuned models across multiple benchmarks. Also, our proposed F-CLIPScore shows enhanced performance to capture global misalignments.
While showing effectiveness in capturing various misalignment types, our analysis reveals that our method inherits weaknesses observed in CLIP. Further examination is needed to improve the detection of misalignments using CLIP variants specifically trained to address these shortcomings.

%

\clearpage

\section{Acknowledgements}

We thank Byungju Han for the discussion and Donghyun Kim for proofreading. We also thank the Vision Understanding team for supporting our work and project namuniv.

\bibliography{aaai25}
\clearpage
\begin{table*}[t]
\small
\centering
\setlength{\tabcolsep}{0.4\tabcolsep}
\begin{tabular}{c|cc|c|c|cccc|cccc}
\hline
Benchmark  & Source& Misalign   & \# samples & Misaligned sentence & 
\multicolumn{4}{|c|}{\# all words } & \multicolumn{4}{|c}{\# misaligned words } \\
 & &domain & & ratio & min & mean & med. & max & min & mean & med. & max\\
\hline
\hline
FOIL & COCO caption & synthetic & 198960 & 0.5 & 6 & 10.67 & 10 & 50 &  1 & 1.0 & 1 & 1 \\
\hline
nocaps-FOIL & Open Images / nocaps & synthetic &  5000 & 0.5 & 7 & 11.53& 11 & 40 & 1 & 1.12 & 1 & 5 \\
\hline
HAT & COCO caption & generated &  400 & 0.34 & 6 & 13.58 & 11 & 35 & 1 & 1.99 & 1 & 12  \\
\hline
\multirow{4}{*}{SeeTRUE-Feedback} &  COCO-con & synthetic &  713 & 1.0 & 6 & 10.31 & 10 & 29 & 1 & 2.49 & 2 & 11  \\
 & COCO-T2I &  generated & 256 & 1.0  & 8 & 10.47 & 10 & 18 & 1 & 4.72 & 4 & 18\\
& Drawbench & generated & 404 & 1.0  & 1 & 9.11 & 8 & 36 & 1 & 4.19 & 4 & 17\\
& Pick-a-pic-con & synthetic & 623 & 1.0  & 4 & 12.10 & 12 & 83 & 1 & 3.15 & 2 & 14\\
\hline
Rich-HF & Pick-a-pic & generated &  955 & 0.74 & 1 & 12.78 & 8 & 182 & 1 & 3.50 & 2 & 100 \\
\hline
\end{tabular}
\caption{\textbf{Detailed statistics of dense misalignment detection benchmark datasets.} The number of words is measured based on white spaces. Misalign domain refers to how misaligned words occur; synthetic refers to that misalignment is synthetically generated(e.g., altering noun by rule), and generated refers to that misalignment is generated by the model (e.g., hallucinated objects in the captioning model).
}
\label{tab:benchmarks-supp}
\end{table*}

\section{A. Further Details of Dense Misalignment Detection Benchmarks}

We provide detailed information about the datasets, as shown in~\cref{tab:benchmarks-supp}.
The FOIL~\cite{shekhar2017foil} and nocaps-FOIL~\cite{petryk2024aloha} dataset consist of misaligned sentences and aligned sentences in a $1$:$1$ ratio. FOIL is characterized by each misaligned sentence containing a single incorrect word, whereas the nocaps-FOIL dataset primarily features sentences with one misaligned word but occasionally includes sentences with multiple misaligned words.
The HAT dataset~\cite{petryk2024aloha} contains a lower proportion of misaligned sentences. In contrast, the SeeTRUE-Feedback dataset~\cite{gordon2023mismatch} exclusively features misaligned sentences, often involving a large number of incorrect words. The Rich-HF dataset displays a wide variety of words, from $1$ to $182$. 

\begin{figure}[h]
     \centering
     \includegraphics[width=0.8\linewidth]{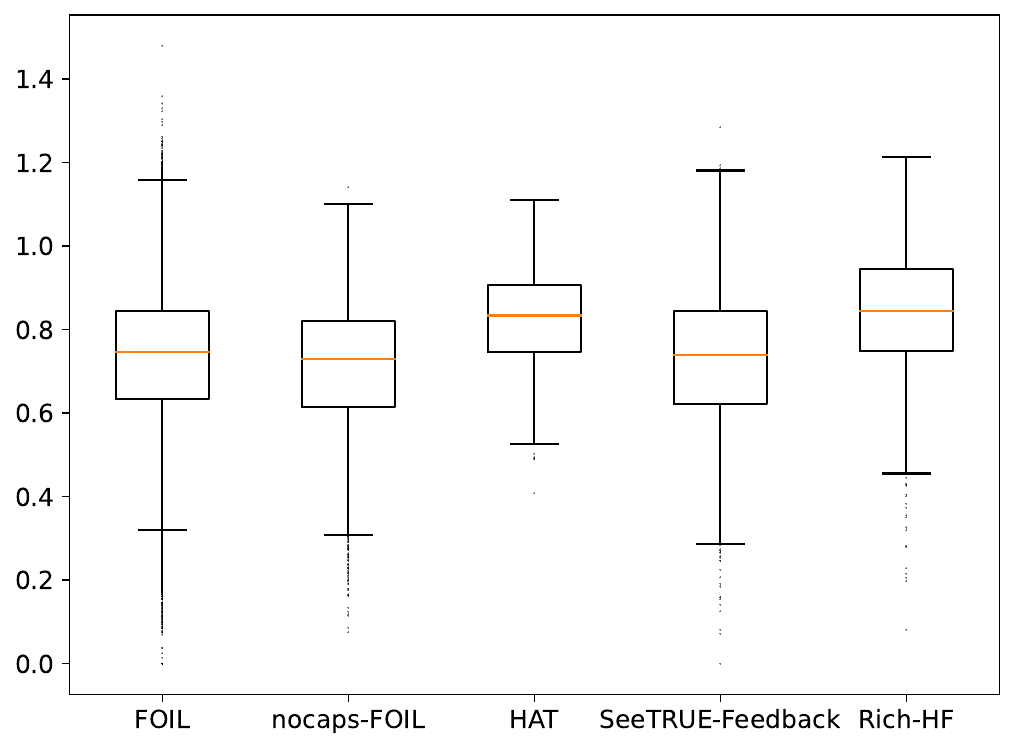}
     \caption{\textbf{Box plots of CLIPScore for the dense misalignment detection benchmark datasets.} We measure scores using ViT-H/$14$ trained on LAION-$2$B~\cite{schuhmann2022laion}.}
     \label{fig:boxplot}
     \vspace{-4mm}
\end{figure}

As shown in~\cref{fig:boxplot}, we demonstrate the distribution of CLIPScore for five benchmarks through box plots. 
In general, we observe the CLIPScore distributions over $0.3$.
The median values of CLIPScore are generally similar across all benchmarks, around $0.8$.
Among the benchmarks, the HAT and Rich-HF show higher median and min values of CLIPScore, which can be attributed to their primary composition of generated image-text pairs.
In contrast, synthetically generated misalignment benchmarks have a relatively wide range of CLIPScore.

\section{B. Additional Qualitative examples}
\subsubsection{FOIL and nocaps-FOIL.}
\begin{figure*}[t!]
     \centering
     \includegraphics[width=1.0\linewidth]{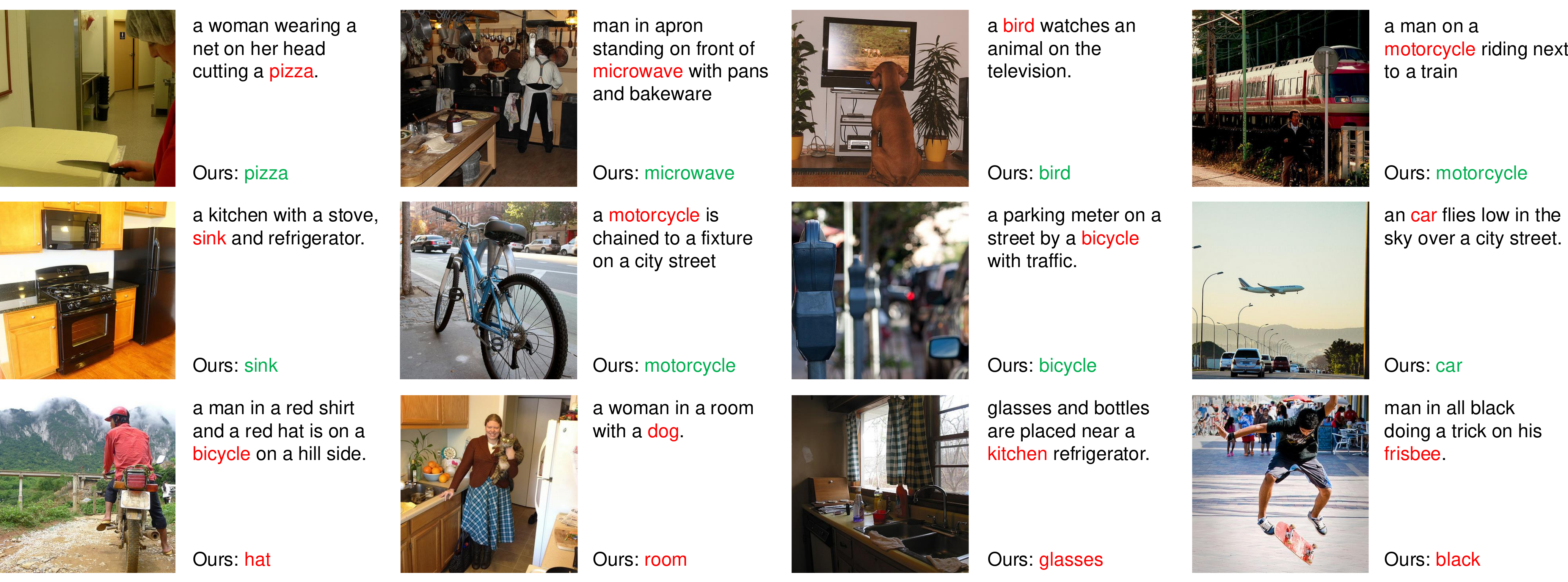}
     \caption{\textbf{Qualitative examples on FOIL.}}
     \label{fig:foil-qual}
     \vspace{-4mm}
\end{figure*}

\begin{figure*}[t!]
     \centering
     \includegraphics[width=1.0\linewidth]{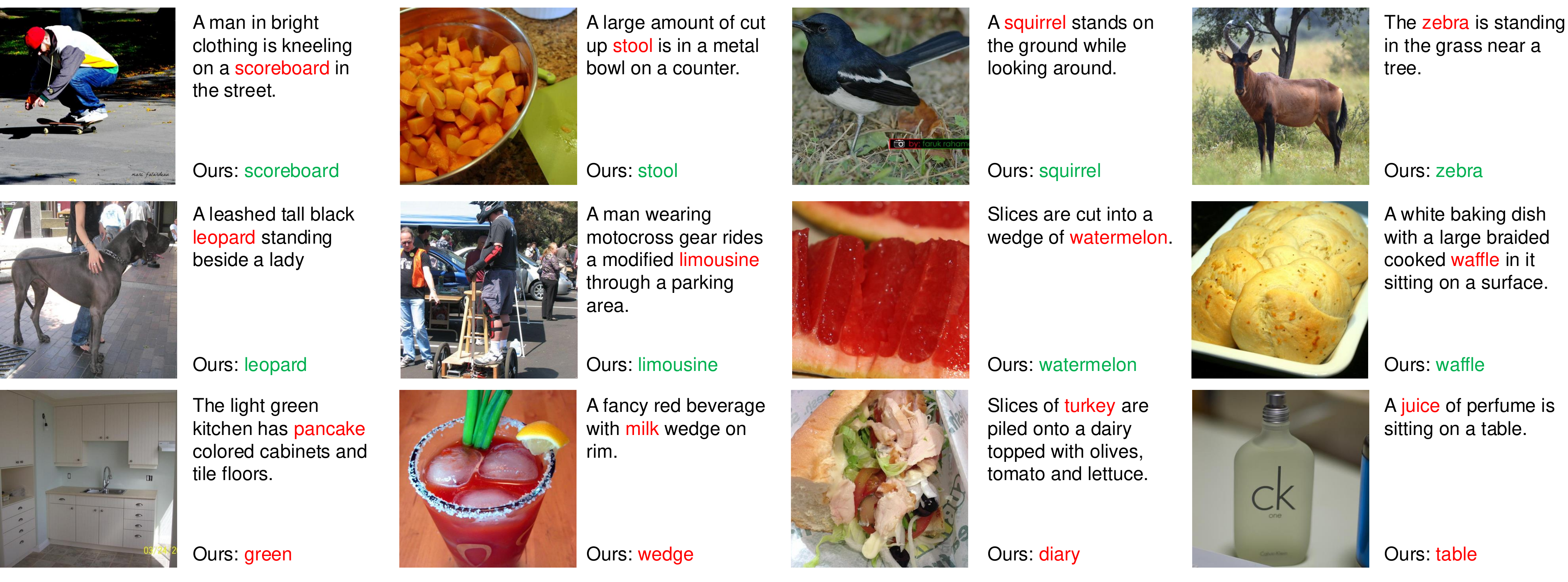}
     \caption{\textbf{Qualitative examples on nocaps-FOIL.}}
     \label{fig:nocaps-foil}
     \vspace{-4mm}
\end{figure*}

As shown in~\cref{fig:foil-qual} and~\cref{fig:nocaps-foil}, qualitative analysis reveals that our method effectively discriminates between conceptually similar but visually distinct objects, such as ``motorcycles'' and ``bicycles'', as well as ``waffles'' and ``bread''. However, it demonstrates reduced accuracy for small objects (e.g., ``glasses'', ``hat'') and background elements (e.g., ``table'', ``room'').  

\subsubsection{HAT.}

\begin{figure*}[t!]
     \centering
     \includegraphics[width=1.0\linewidth]{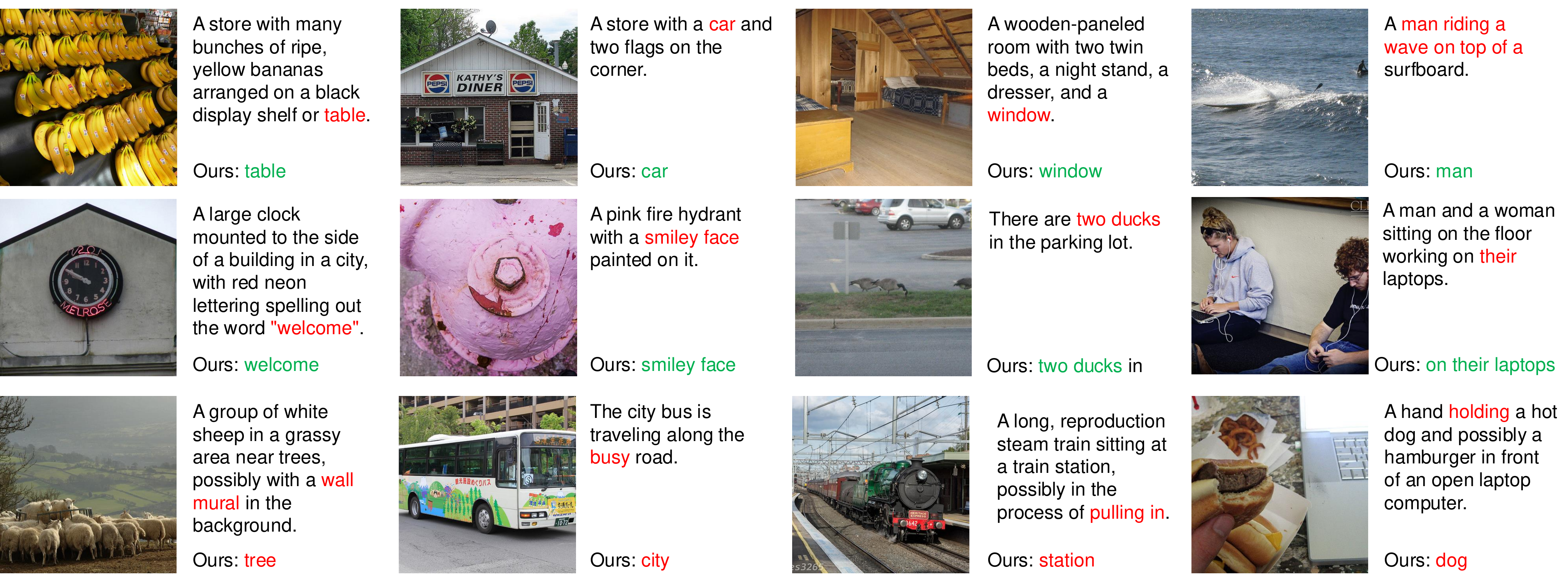}
     \caption{\textbf{Qualitative examples on HAT.}}
     \label{fig:hat}
     \vspace{-4mm}
\end{figure*}

Additional examples from the HAT dataset are presented in~\cref{fig:hat}. Our method demonstrates proficiency in identifying a wide spectrum of misalignments, including numbers (e.g., ``two ducks'') and abstract (e.g., ``smiley face''). However, the method's performance is limited when detecting misalignments in adjectives (e.g., ``busy'') and action verbs (e.g., ``pulling in'', ``holding'').

\begin{figure*}[t!]
     \centering
     \includegraphics[width=1.0\linewidth]{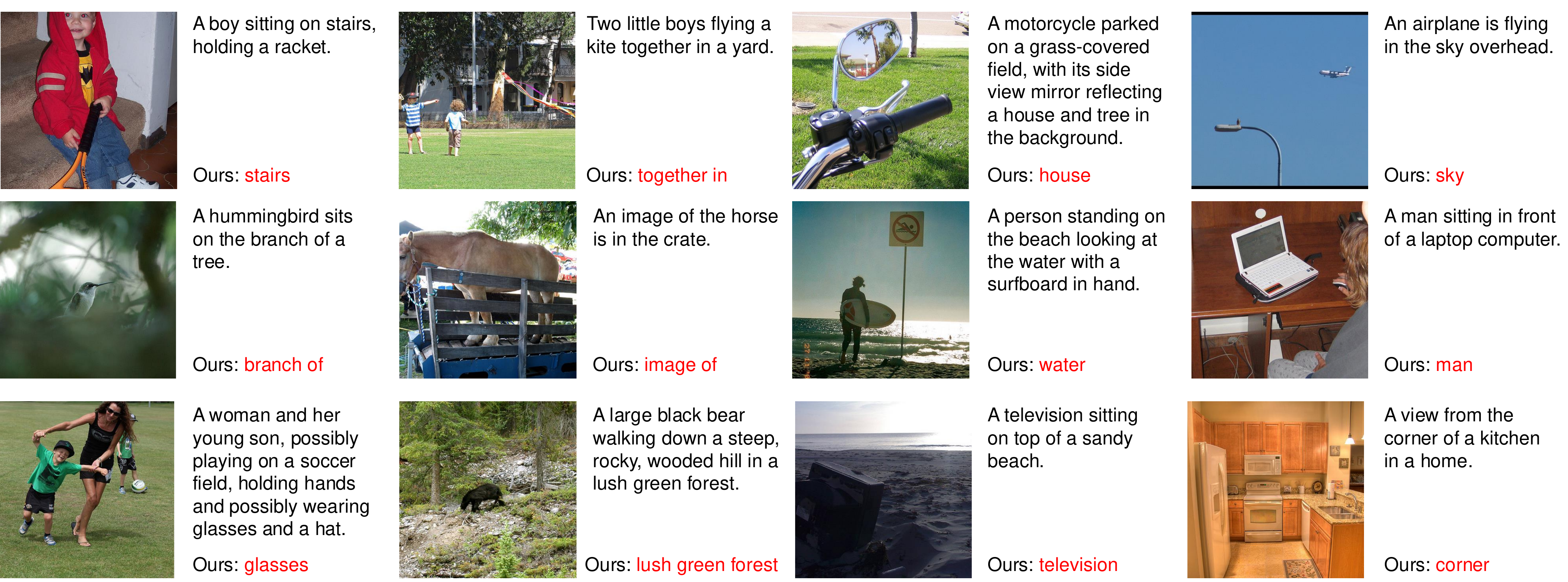}
     \caption{\textbf{False positive cases on HAT.}}
     \label{fig:hat2}
     \vspace{-4mm}
\end{figure*}

We further investigate the factors contributing to the decreased Average Precision (AP) on the HAT dataset in~\cref{fig:hat2}. Our analysis reveals that false positives predominantly occur in scenarios involving background elements (e.g., ``stairs'', ``sky''), small or indistinct objects (e.g., ``house'', ``television''), and descriptive phrases (e.g., ``in the corner'', ``image of''). This pattern of errors aligns with the previously noted tendency of CLIP to prioritize foreground elements, suggesting that our method inherits CLIP's bias towards salient visual features at the expense of more subtle or contextual information.

\subsubsection{SeeTRUE-Feedback.}
\begin{figure*}[t!]
     \centering
     \includegraphics[width=1.0\linewidth]{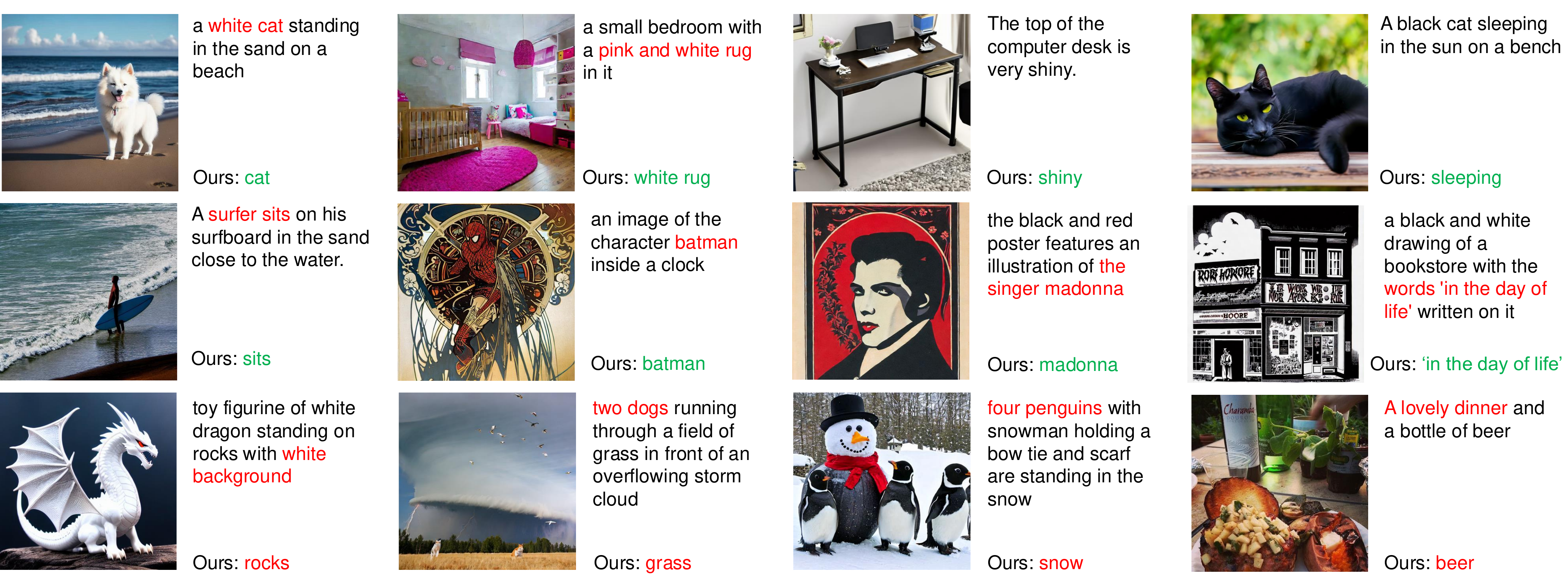}
     \caption{\textbf{Qualitative examples on SeeTRUE-Feedback.}}
     \label{fig:seetrue}
     \vspace{-4mm}
\end{figure*}

Qualitative evaluations of our method on the SeeTRUE-Feedback dataset are presented in~\cref{fig:seetrue}. The results demonstrate enhanced capability in identifying diverse forms of misalignment, including objects, colors, attributes (e.g., ``shiny''), and some actions (e.g., ``sits'', ``sleeping''). Notably, the method exhibits proficiency in recognizing entity-level objects (e.g., ``batman'', ``madonna''), which can not be easily captured by existing approaches. In line with known CLIP capabilities, our approach also demonstrates some optical character recognition (OCR) ability~\cite{radford2021learning,lin2025parrot}, as evidenced by samples such as ``in the day of life''. However, persistent limitations in detecting background elements and small objects underscore areas for future improvement.

\begin{figure*}[t!]
     \centering
     \includegraphics[width=1.0\linewidth]{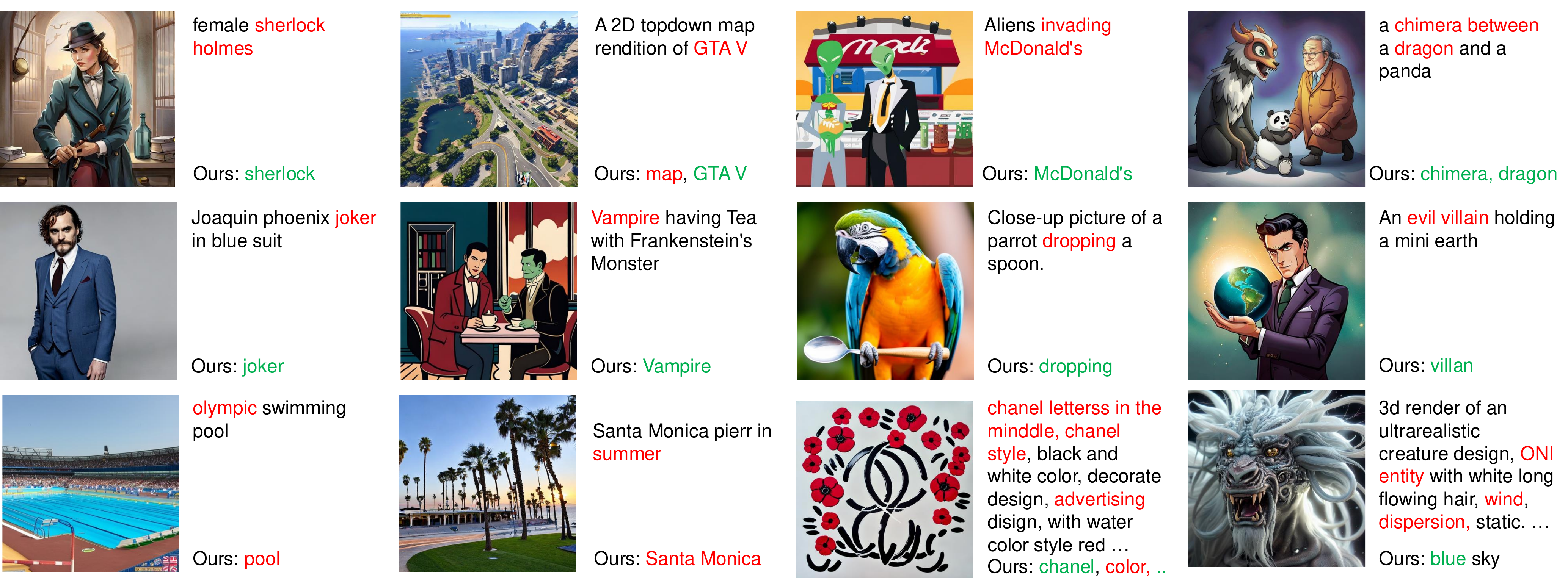}
     \caption{\textbf{Qualitative examples on Rich-HF.}}
     \label{fig:rich}
     \vspace{-4mm}
\end{figure*}

\begin{figure}[t!]
    \centering
    \begin{minipage}{0.5\textwidth}
        \begin{subfigure}{\linewidth}
            \includegraphics[width=\textwidth]{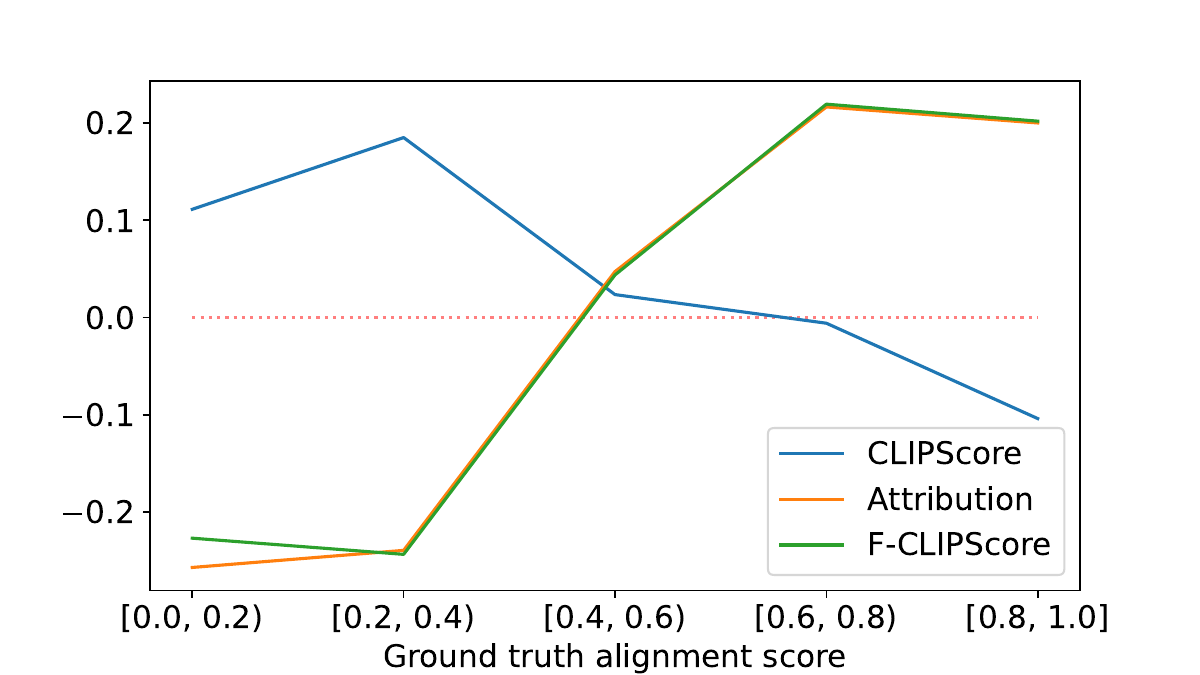}
            \caption{Pearson correlation coefficients}
            \label{fig:pearson}
        \end{subfigure}
    \end{minipage}
    \begin{minipage}{0.5\textwidth}
        \begin{subfigure}{\linewidth}
            \includegraphics[width=\textwidth]{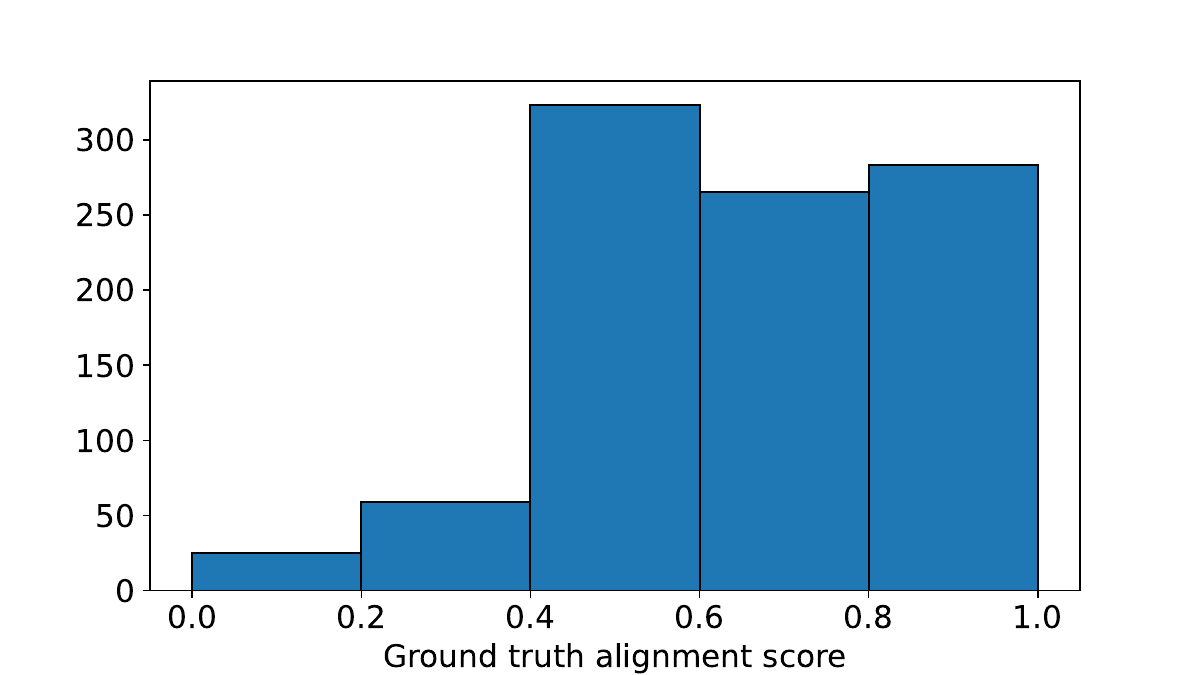}
            \caption{Histogram}
            \label{fig:hist}
        \end{subfigure}
    \end{minipage}
\caption{\textbf{Analysis of group-wise Pearson correlation coefficients and histogram distributions of ground truth alignment scores in Rich-HF dataset.} We conduct analysis by dividing the ground truth alignment scores into five groups.}
\label{fig:richhf_fclip_ablation}
\vspace{-4mm}
\end{figure}

\subsubsection{Rich-HF.}
Qualitative examples in Rich-HF are represented on~\cref{fig:rich}. Rich-HF dataset, which consists of real user prompts, reveals our method's particular strength in identifying misalignments related to well-known entities, such as ``GTA V'', ``Sherlock'', ``Chimera''), while missing some entities (e.g., ``ONI'', ``Santa Monica''). The Rich-HF dataset is characterized by a high proportion of highly descriptive and style-related terms (e.g., ``advertising design''), which often present challenges for CLIP-based models to accurately capture and evaluate. Also, our model's architecture constrains input to the first 77 tokens, resulting in limited performance on prompts with a significantly large number of words.

\section{C. Further Results on Challenging Benchmarks}

As research on CLIP progresses, challenging benchmarks~\cite{tong2024eyes, hsieh2023sugarcrepe, yuksekgonul2023and, ma2023crepe} addressing its weaknesses have been proposed. They are characterized by hard negatives with various visual and textual patterns, including attribution, relation, and order. Therefore, we compare CLIPScore and F-CLIPScore using the latest MMVP~\cite{tong2024eyes} and SugarCrepe~\cite{hsieh2023sugarcrepe} benchmarks.
MMVP is composed of image and text pairs that include nine visual patterns that CLIP particularly struggles with, and SugarCrepe consists of hard negatives constructed using add, replace, and swap methods for attributes, relations, and objects.
For experiments, we utilize the OpenAI ViT-L/$14$ variant, following the configuration used in MMVP.

\begin{table*}[t!]
\small
\centering
    \setlength{\tabcolsep}{0.5\tabcolsep}
    \begin{center}
        \begin{tabular*}{\linewidth}{@{\extracolsep{\fill}} l|ccccccccc|c}
        \hline
            Model & \makecell{Orientation \\ Direction} & \makecell{Presence} & \makecell{State\\Condition} & \makecell{Quantity\\Count} & \makecell{Positional\\Relational} & \makecell{Color\\ Appearance} & \makecell{Structural\\Physical} & Texts & \makecell{Viewpoint\\Perspective} & \makecell{MMVP\\Average}\\
            \hline
            CLIPScore & 6.7 & 13.3 & 20.0 & \textbf{13.3} & \textbf{6.7} & \textbf{53.3} & \textbf{26.7} & \textbf{6.7} & 13.3 & 17.8 \\
            F-CLIPScore & \textbf{33.3} & \textbf{26.7} & \textbf{40.0} & \textbf{13.3} & 0.0 & 40.0 & \textbf{26.7} & \textbf{6.7} & \textbf{33.3} & \textbf{24.4}\\
            \hline
        \end{tabular*}
    \end{center}
    \vspace{-1.0em}
    \caption{\textbf{Experiment results on MMVP benchmark.} The scores of CLIPScore are reproduced using the publicly available repository. MMVP Average represents the macro average of the scores across nine visual patterns.}
    \vspace{-2.mm}
    \label{table:mmvp}
\end{table*}

\begin{table*}[t!]
\small
\centering
    \setlength{\tabcolsep}{0.5\tabcolsep}
    \begin{center}
        \begin{tabular*}{\linewidth}{@{\extracolsep{\fill}} l|cccccccc}
        \hline
            \multirow{2}{*}{Model} & \multicolumn{3}{c}{REPLACE} & \multicolumn{2}{c}{SWAP} & \multicolumn{2}{c}{ADD} & \multirow{2}{*}{Overall}\\
            \cline{2-4} \cline{5-6} \cline{7-8}  & Object & Attribute & Relation & Object & Attribute & Object & Attribute \\
            \hline
            CLIPScore & \textbf{94.07} & \textbf{79.19} & \textbf{65.15} & \textbf{60.41} & \textbf{62.31} & 78.32 & 71.53 & 76.78 \\
            F-CLIPScore & 92.07 & 75.38 & 59.89 & 57.96 & 58.26 & \textbf{88.94} & \textbf{84.25} & \textbf{78.60}\\
            \hline
        \end{tabular*}
    \end{center}
    \vspace{-1.0em}
    \caption{\textbf{Experiment results on SugarCrepe benchmark.} The scores of CLIPScore are reproduced using the publicly available repository. Overall denotes the micro average score across the entire dataset.}
    \vspace{-2.mm}
    \label{table:sugarcrepe}
\end{table*}

F-CLIPScore demonstrates substantial improvements on the MMVP dataset, with an overall gain of over $6.6$\%p as shown in~\cref{table:mmvp}. 
Notably, we observe significant enhancements in detecting misalignments related to orientation \& direction, viewpoint \& perspective, and state \& condition.
These results underscore our method's effectiveness in distinguishing subtle misalignments.
On the SugarCrepe benchmark, our approach achieves an overall improvement of $1.8$\%p gain as shown in~\cref{table:sugarcrepe}.
The main performance gain comes from the ``add'' form of hard negatives with a $10.6$\%p improvement in adding object concept and $12.7$\%p improvement in adding attribute concept. We hypothesize that adding words to a sentence can introduce a bias in CLIPScore due to its sensitivity to length (i.e., a longer sentence tends to yield higher CLIPScore regardless of correctness). In contrast, F-CLIPScore is more sensitive to detailed misalignments, improving performance.
However, F-CLIPScore shows decreased performance compared to CLIPScore for ``replace'' and ``swap'' forms of hard negatives.
We provide further analysis of the different tendencies of CLIPScore and F-CLIPScore in the following section.

\section{D. Further Analyses on F-CLIPScore}
To evaluate the efficacy of F-CLIPScore, we present qualitative results on the nocaps-FOIL dataset. To elucidate the divergent tendencies between CLIPScore and F-CLIPScore, we sort the scores both in ascending and descending order. We then identify examples where the rank differential between the two metrics falls within the top 15\% in opposite directions (e.g., instances where CLIPScore ranks in the top 1\% while F-CLIPScore ranks in the bottom 13\%).

\begin{figure}[t]
     \centering
     \includegraphics[width=0.9\linewidth]{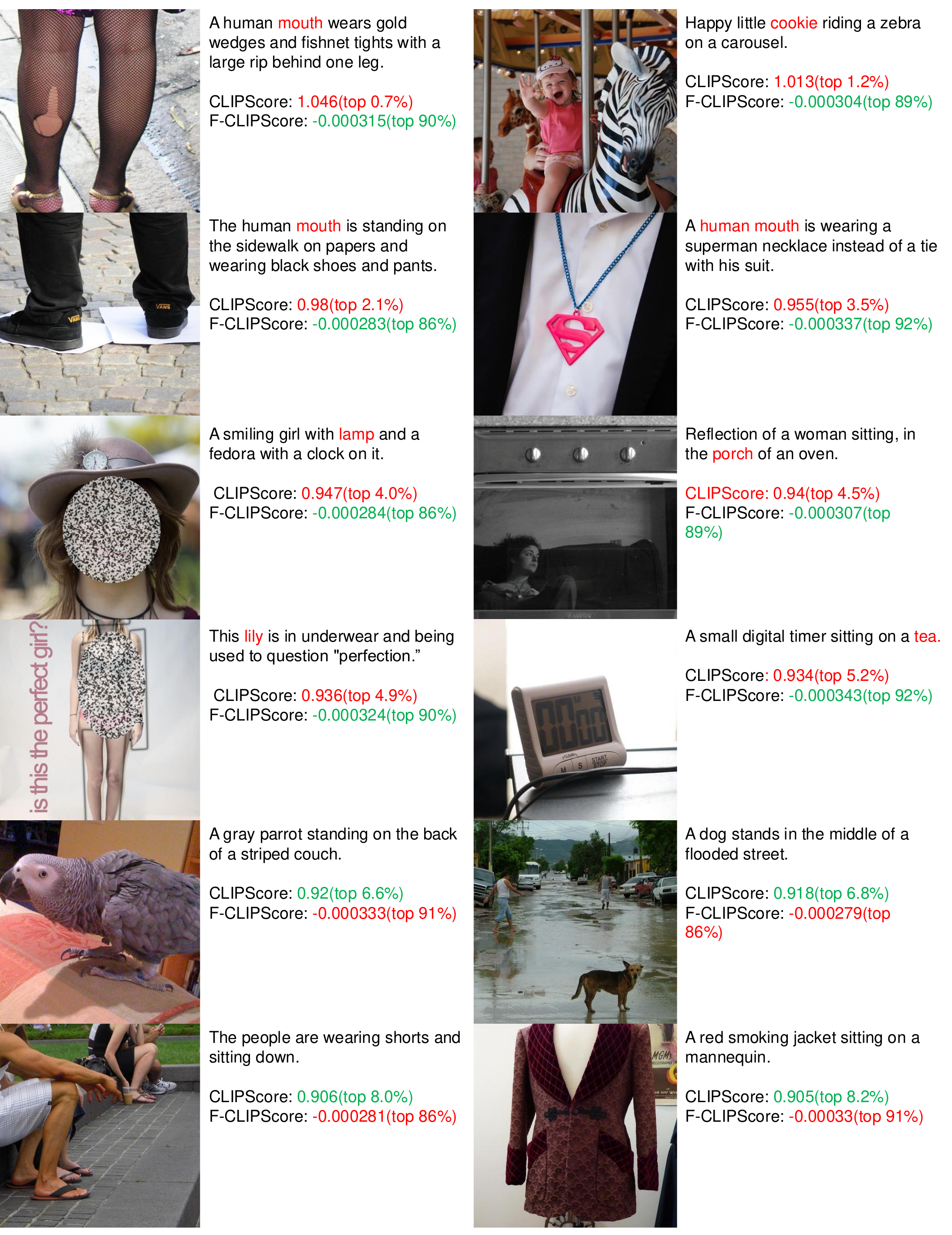}
     \caption{\textbf{Qualitative examples on nocaps-FOIL sorted by CLIPScore in descending order.}}
     \label{fig:fclip1}
     \vspace{-4mm}
\end{figure}

\begin{figure}[t]
     \centering
     \includegraphics[width=0.9\linewidth]{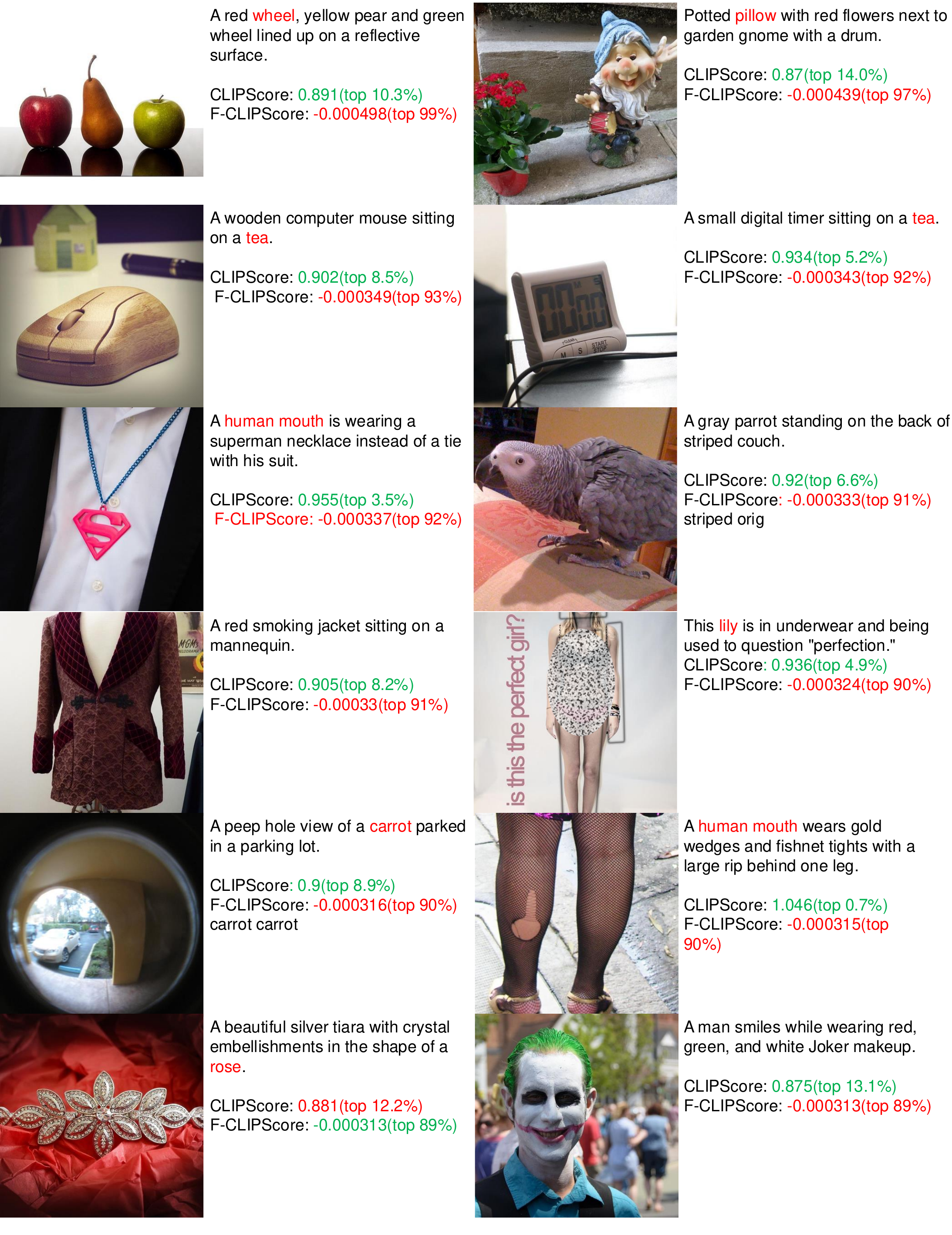}
     \caption{\textbf{Qualitative examples on nocaps-FOIL sorted by F-CLIPScore in ascending order.}}
     \label{fig:fclip3}
     \vspace{-4mm}
\end{figure}

\cref{fig:fclip1} and~\cref{fig:fclip3} illustrate examples where CLIPScore is high and F-CLIPScore is low. The results show that CLIPScore tends to assign significantly high values when salient and specific terms are correctly matched (e.g., ``fishnet tights'', ``zebra'', ``superman''), even in the presence of obvious misalignments (e.g., ``cookie'' instead of ``child'') in the caption. In contrast, F-CLIPScore exhibits greater sensitivity to these misaligned elements. This suggests that CLIPScore may be disproportionately influenced by the presence of correctly identified prominent features, while F-CLIPScore is sensitive to misaligned words.

\begin{figure}[t]
     \centering
     \includegraphics[width=0.9\linewidth]{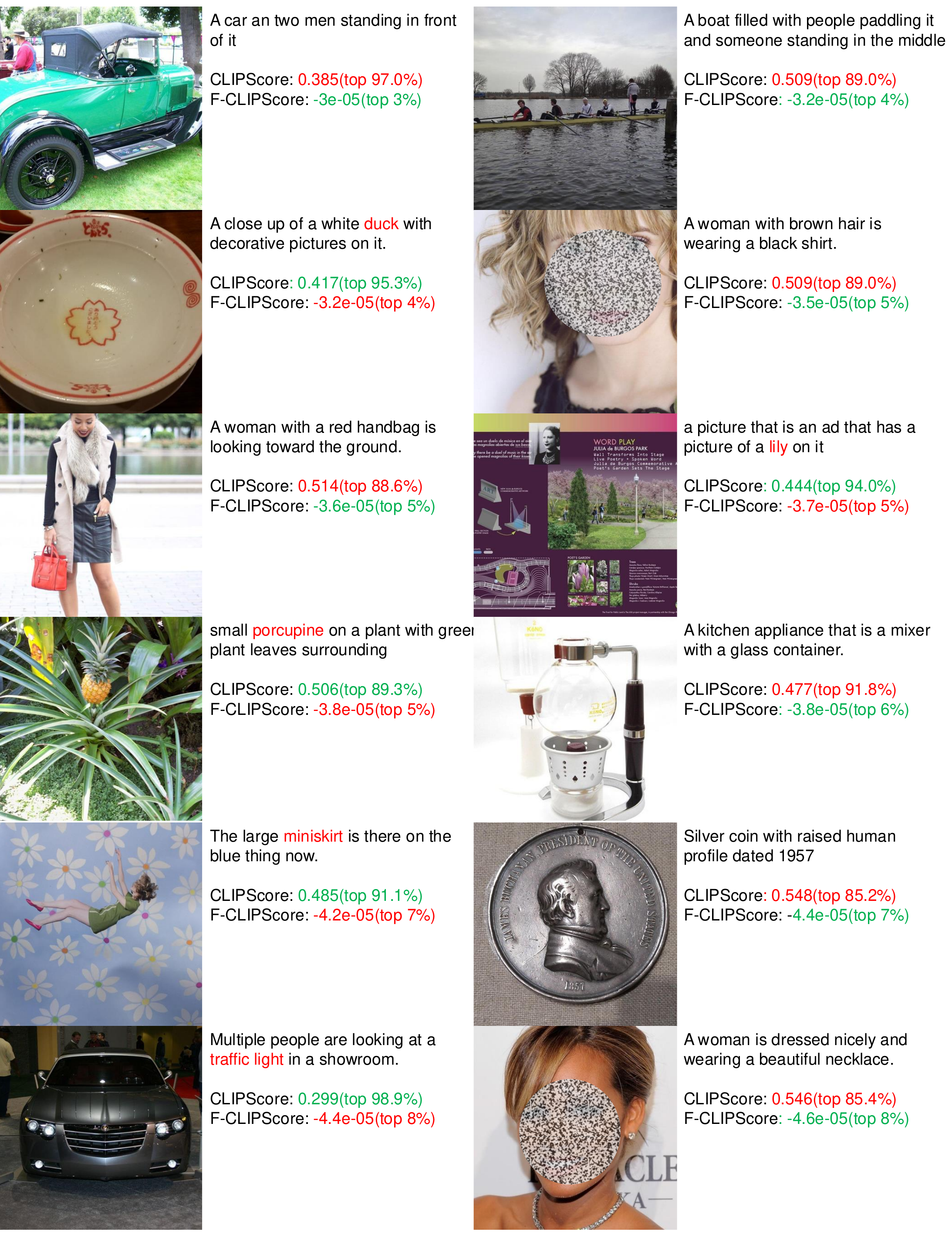}
     \caption{\textbf{Qualitative examples on nocaps-FOIL sorted by F-CLIPScore in descending order.}}
     \label{fig:fclip4}
     \vspace{-4mm}
\end{figure}

\begin{figure}[t]
     \centering
     \includegraphics[width=0.9\linewidth]{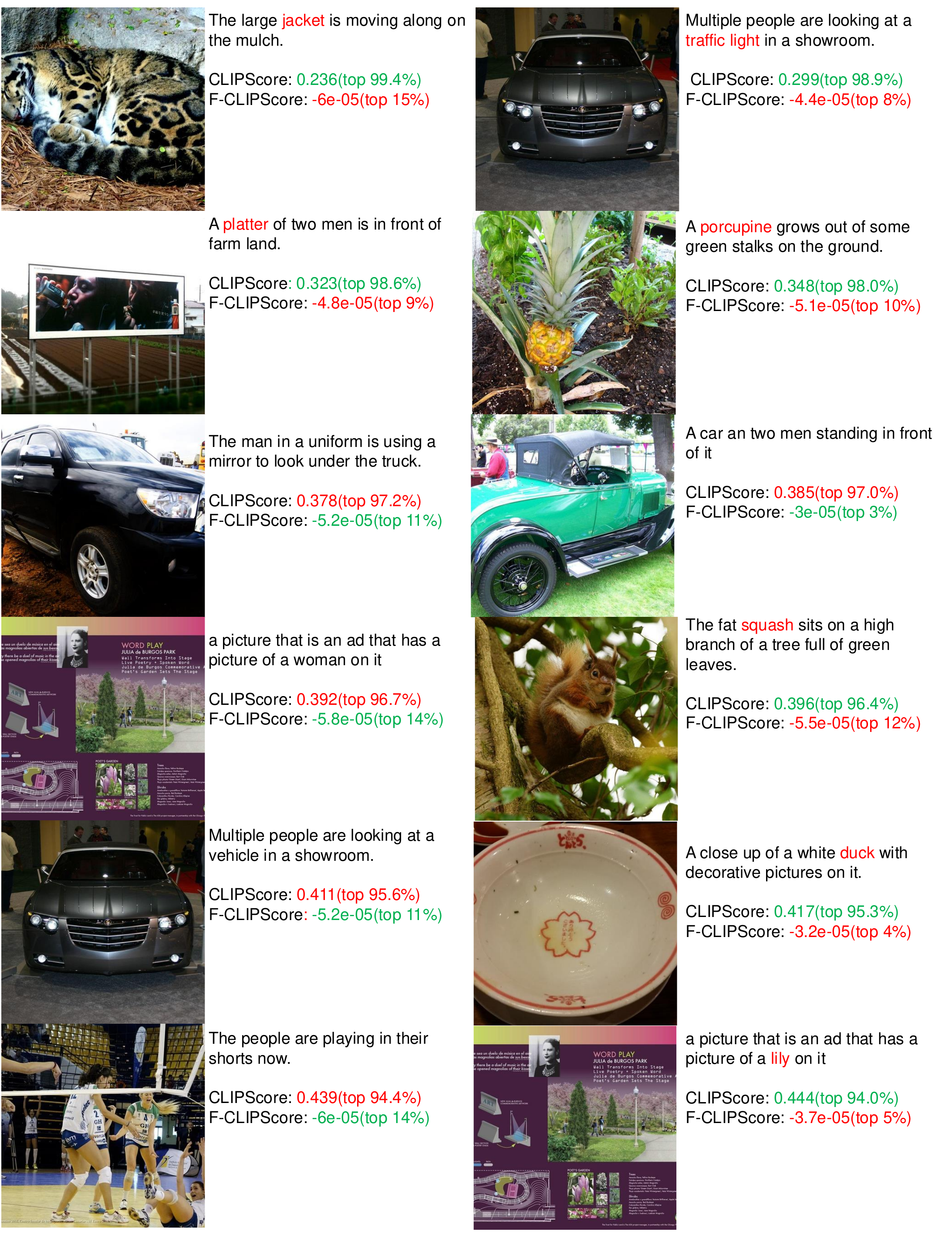}
     \caption{\textbf{Qualitative examples on nocaps-FOIL sorted by CLIPScore in ascending order.}}
     \label{fig:fclip2}
     \vspace{-4mm}
\end{figure}

\cref{fig:fclip4} and~\cref{fig:fclip2} show examples where F-CLIPScore is high, but CLIPScore is low. The observed pattern in these examples is subtle, revealing a complex relationship between the two metrics. 
In general, when captions are perfectly aligned but composed words are mainly common (e.g., ``man'', ``car'', ``woman''), CLIPScore reports low value. It corresponds with a study that reveals CLIP has a bias for specified words~\cite{pezzelle2023dealing}. In contrast, F-CLIPScore reports high values for those captions. 
However, when apparent misalignments (e.g., ``duck'') occur, F-CLIPScore unexpectedly yields high values in these cases. 
We observe that when CLIPScore is extremely low, the gradients are distributed across multiple tokens, resulting in few tokens having gradients lower than epsilon.
This phenomenon highlights a limitation of the proposed F-CLIPScore metric, necessitating careful interpretation, especially when CLIPScore is low. However, as discussed in~\cref{fig:boxplot}, the median CLIPScore typically ranges between $0.6$ and $0.8$, indicating that such extremely low scores are not common in the generated output. It may be advisable to exclude outlier samples whose CLIPScore is extremely low and apply F-CLIPScore selectively to capture factual alignments.

From the results, we conduct further analysis of the relationship between CLIPScore and F-CLIPScore.
Since the Rich-HF dataset contains human-labeled alignment scores for images and text, we divide them into five groups and obtain Pearson correlation coefficients with CLIPScore, attribution scores, and F-CLIPScore for each group, as shown in~\cref{fig:richhf_fclip_ablation}.
Note that attribution scores indicate $\sum_j\text{mis}(w_j) \cdot w_j$ on the Equation ($11$) from the main manuscript.
As shown in~\cref{fig:pearson}, CLIPScore and F-CLIPScore demonstrate opposite patterns: CLIPScore and F-CLIPScore show low performance in groups with high and low ground truth alignment scores, respectively.

As mentioned in the previous paragraph, because of this tendency, CLIPScore and F-CLIPScore should be interpreted carefully. 
However, since samples with low alignment scores are generally not generated in recent generative models as shown in~\cref{fig:hist}, F-CLIPScore shows better performance than CLIPScore and will have higher usability for detecting misalignments.

\section{E. Ablation on Backbones}

\begin{table}[t!]
\footnotesize
\centering
    \setlength{\tabcolsep}{0.3\tabcolsep}
    \begin{center}
        \begin{tabular*}{\linewidth}{@{\extracolsep{\fill}} ccc|c|cc}
            \hline
            backbone & source & pretrained &IN acc. & LA & AP \\
            \hline
            ViT-B/32 & openai  & WIT-400M & 0.632 & 0.602 & 0.723   \\
            ViT-B/32 & openclip& LAION-2B & 0.656 &0.667 & 0.760 \\
            ViT-B/16 & openai  & WIT-400M & 0.687 & 0.679 & 0.747   \\
            ViT-L/14 & openai& WIT-400M & 0.753 &0.653 & 0.781 \\
            ViT-L/14 & openclip & LAION-2B & 0.753 &0.728 & 0.796\\
            ViT-H/14 & openclip& LAION-2B &0.780 &0.716& 0.806 \\
            ViT-g/14& openclip & LAION-2B &0.766 &0.706& 0.806 \\
            \hline
        \end{tabular*}
    \end{center}
    \vspace{-1.5em}
    \caption{\textbf{Various backbone comparisons on nocaps-FOIL dataset}. IN acc. refers to ImageNet accuracy.} 
    \vspace{-4.mm}
    \label{table:backbone}
\end{table}

We conduct an ablation study examining the impact of varying model backbone and pretraining corpora on performances~\cite{radford2021learning,cherti2023reproducible}. Experiments were performed on the nocaps-FOIL dataset, with $\tilde{l}$ set to the last three layers of each text encoder. Results in~\cref{table:backbone} demonstrate a scaling law in general. Performance in Localization Accuracy (LA) and Average Precision (AP) improve with increased backbone size. Notably, ImageNet accuracy shows a positive correlation with both LA and AP, indicating a strong dependence on CLIP backbone performance. Models pre-trained on LAION-2B~\cite{castro2023scalable} consistently outperform those trained on 400M WebImageText (WIT)~\cite{radford2021learning} in both metrics. For instance, two ViT-L/14 variants, both achieving identical ImageNet accuracy~\cite{deng2009imagenet} but pre-trained on different datasets (WIT-400M vs. LAION-2B), show varied performances. The variant trained on LAION-2B shows significantly improved performance, showing superior capabilities in detecting a wider range of misaligned words.

\section{F. Discussions}

Despite the efficacy of our method, several limitations merit further examination. Firstly, as discussed in our analysis, our approach inherits known weaknesses from CLIP. However, various studies have addressed these shortcomings of CLIP, focusing on issues such as numbers~\cite{paiss2023teaching}, compositionality~\cite{yuksekgonul2023and}, or small or insalient objects~\cite{yaofilip,mukhoti2023open}.
We leave evaluating CLIP variants specifically trained to address these shortcomings as future work. 

Secondly, the token length constraint poses challenges for long-context understanding. Although CLIP models trained on extended sequences~\cite{zhang2025long} may partially alleviate this issue, it remains a potential drawback for tasks requiring broader contextual comprehension. 

Thirdly, our analysis reveals that F-CLIPScore performs poorly when CLIPScore is extremely low due to the distribution of gradients.
While our research primarily focuses on detecting dense misalignments from generated models, which typically have relatively well-aligned image-caption pairs, applying F-CLIPScore to noisy alt-text could be suboptimal. A sophisticatedly designed F-CLIPScore that can adapt to significantly low CLIPScores would be beneficial.

Lastly, our method generally shows higher recall but lower precision. It would be an interesting area of research to further sophisticate the design of the attribution calculation or refinement method to achieve more precise results. Despite its low precision, it is worth noting that its superior computational efficiency facilitates large-scale applications. We leave it as future work to explore whether leveraging this scalability for extensive data correction (e.g., removing misaligned words in captions) or for large-scale reinforcement learning from AI feedback could potentially enhance overall performance.

\clearpage

\clearpage

\end{document}